\documentclass{article} 
\usepackage{iclr2024_conference,times}

\usepackage{amsmath,amsfonts,bm}

\def\eqref#1{equation~\ref{#1}}

\def\1{\bm{1}}

\DeclareMathAlphabet{\mathsfit}{\encodingdefault}{\sfdefault}{m}{sl}
\SetMathAlphabet{\mathsfit}{bold}{\encodingdefault}{\sfdefault}{bx}{n}

\usepackage{hyperref}
\usepackage{url}
\usepackage{multirow}
\usepackage{graphicx}
\usepackage{algorithm}
\usepackage{soul}
\usepackage{ulem}
\usepackage{epstopdf}
\usepackage{subfig}
\usepackage{epsfig}
\usepackage{wrapfig}
\usepackage{amsmath}

\usepackage{adjustbox}

\usepackage{algpseudocode}
\usepackage[noend]{algcompatible}
\usepackage{algorithmicx}

\usepackage{float}
\usepackage{colortbl} 
\usepackage{xcolor}  
\definecolor{mypink}{rgb}{1,0.928,0.928}
\definecolor{mydarkpink}{rgb}{1,0.7,0.7}
\definecolor{lightblue}{RGB}{57, 139, 250} 

\usepackage[utf8]{inputenc}
\usepackage{booktabs} 

\usepackage{tikz}
\usepackage{arydshln}

\usepackage{xcolor}

\title{FedLoGe: Joint Local and Generic Federated Learning under Long-tailed Data}
\author{%
\textbf{Zikai Xiao}\textsuperscript{1\thanks{Co-first author.}}, 
\textbf{Zihan Chen}\textsuperscript{2\footnotemark[1]},
\textbf{Liyinglan Liu}\textsuperscript{3}, 
\textbf{Yang Feng}\textsuperscript{4},
\textbf{Jian Wu}\textsuperscript{1},
\textbf{Wanlu Liu}\textsuperscript{1},\\
\textbf{Joey Tianyi Zhou}\textsuperscript{5,6},
\textbf{Howard Hao Yang}\textsuperscript{1},
\textbf{Zuozhu Liu}\textsuperscript{1\thanks{Corresponding author.}} \\
\textsuperscript{1}Zhejiang University,\\
\textsuperscript{2}Singapore University of Technology and Design,\\
\textsuperscript{3}University of Electronic Science and Technology of China, \\
\textsuperscript{4}Angelalign Technology Inc,\\
\textsuperscript{5}IHPC, Agency for Science, Technology and Research, Singapore,\\
\textsuperscript{6}CFAR, Agency for Science, Technology and Research, Singapore\\
\texttt{\href{mailto:zikai@zju.edu.cn}{\texttt{zikai@zju.edu.cn}}}
}
  
\iclrfinalcopy 
\begin{document}

\maketitle

\algnewcommand\algorithmicreturn{\textbf{return} }
\algnewcommand\RETURN{\State \algorithmicreturn}%

\algrenewcommand\algorithmicrequire{\textbf{function}} 

\algnewcommand\algorithmicreq{\textbf{Require:} }
\algnewcommand\REQ{\STATEx \algorithmicreq{}}%

\begin{abstract}
Federated Long-Tailed Learning (Fed-LT), a paradigm wherein data collected from decentralized local clients manifests a globally prevalent long-tailed distribution, has garnered considerable attention in recent times. 
In the context of Fed-LT, existing works have predominantly centered on addressing the data imbalance issue to enhance the efficacy of the generic global model while neglecting the performance at the local level. 
In contrast, conventional Personalized Federated Learning (pFL) techniques are primarily devised to optimize personalized local models under the presumption of a balanced global data distribution. 
This paper introduces an approach termed \textbf{Fed}erated \textbf{Lo}cal and \textbf{Ge}neric Model Training in Fed-LT (FedLoGe), which enhances both local and generic model performance through the integration of representation learning and classifier alignment within a neural collapse framework. 
Our investigation reveals the feasibility of employing a shared backbone as a foundational framework for capturing overarching global trends, while concurrently employing individualized classifiers to encapsulate distinct refinements stemming from each client’s local features.
Building upon this discovery, we establish the Static Sparse Equiangular Tight Frame Classifier (SSE-C), inspired by neural collapse principles that naturally prune extraneous noisy features and foster the acquisition of potent data representations. 
Furthermore, leveraging insights from imbalance neural collapse's classifier norm patterns, we develop Global and Local Adaptive Feature Realignment (GLA-FR) via an auxiliary global classifier and personalized Euclidean norm transfer to align global features with client preferences. Extensive experimental results on CIFAR-10/100-LT, ImageNet-LT, and iNaturalist demonstrate the advantage of our method over state-of-the-art pFL and Fed-LT approaches. Our codes are available at \href{https://github.com/ZackZikaiXiao/FedLoGe}{\textcolor{lightblue}{https://github.com/ZackZikaiXiao/FedLoGe}}.

\end{abstract}

\section{Introduction}
Federated learning (FL) enables collaborative model training across decentralized clients without exposing local private data \citep{mcmahan2017communication, kairouz2021advances}.
Recent work further investigates the federated long-tailed learning (Fed-LT) task, where the global data exhibits long-tailed distributions and local clients hold heterogeneous distributions \citep{chen2022towards, shang2022federated}. They usually learn a well-trained generic global model, whose performance might degrade when universally applied to all clients with diverse data, degrading its practical applicability. For example, in the realm of smart healthcare, as demonstrated by \cite{lee2020federated}, \cite{chen2022personalized}, and  \cite{elbatel2023federated}, the capacity of the global model to deliver high-quality diagnostics is limited, as patient distributions vary across specialized hospitals. Additionally, in cross-institutional financial applications like credit scoring \citep{dastile2020statistical} and fraud detection \citep{awoyemi2017credit}, individuals from different regions or age groups may exhibit dissimilar credit patterns. 

A high-quality global model with balanced performance could speed up local adaptation and attract new clients, while personalized models aim to provide enhanced local performance by considering local data characteristics. Nevertheless, existing works on Fed-LT have primarily focused on addressing the imbalance in the context of the global long-tailed data~\citep{yang2023integrating, qian2023long,xiao2023fedgrab}, neglecting the tailoring of models to the needs of individual clients, since local data statistics could be diverse and not necessarily long-tailed. 
Personalized federated learning (pFL)~\citep{tan2022towards}, which trains customized local models for a single or a group of clients, offers an alternative solution to prioritize each client's (or group's) distinct data statistics and preferences, in which the global generic model is deemed as a bridge for training local personalized models and boosting the local performance with expressive representations~\citep{li2021fedbn, collins2021exploiting, li2021ditto}.
However, conventional pFL approaches are not supposed to attain a superior global generic model in Fed-LT.

\begin{figure}[htbp]
    \centering
    \scriptsize
    \begin{tabular}{cc}
        \includegraphics[width=0.42\columnwidth]{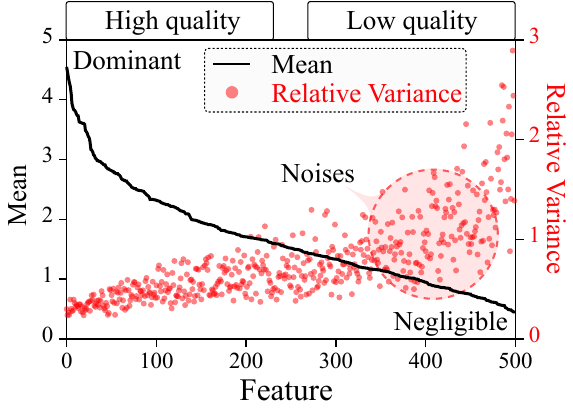}  &
        \includegraphics[width=0.42\columnwidth]{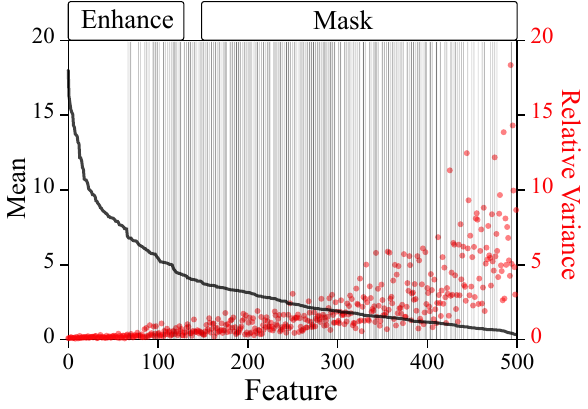}           \\
        (a) & (b)
    \end{tabular}
    \caption{(a): The mean (sorted in descending order) and variance of class means unveil feature degeneration: The feature collapse property ceases to prevail, and features with diminished means \textbf{exhibit substantial variance rather than being zero}; (b): After training the backbone with SSE-C, in which \textbf{noisy features with bigger variance are partially pruned} (gray shaded vertical lines), and \textbf{enhance the quality} (smaller variance) of dominant features.}
    \vspace{-5pt}
    \label{fig:mean_std}
\end{figure}

Designing a framework to \textbf{simultaneously train global and local models} in the presence of Fed-LT remains a critical challenge. Inspired by~\cite{kang2019decoupling}, we find that the generality and transferability of feature extractors are significantly superior to classifiers~\citep{kim2022broad, vasconcelos2022proper}. On the contrary, adjusting the classifier has proven to be quite effective in addressing the imbalance and heterogeneity issues~\citep{li2022long, zhang2022federated}. In other words, the feature extractor can serve as the cornerstone to reflect global trends, while adjusting classifiers can induce the model to adaptively achieve superior personalized performance across the server and heterogeneous clients. Consequently, we conceptualize our model learning with \textbf{two intertwined processes: global representation learning and imbalanced/heterogeneous classifier adaptation}. Adopting this viewpoint, we identify two key challenges for personalized Fed-LT:

\noindent \texttt{C1:} \textit{How to learn effective representations under heterogeneous and imbalanced data?}

Due to heterogeneity, each client captures different feature distributions, leading to divergence during model aggregation and inferior global performance. Recent studies show that training with a fixed classifier can reduce divergence among heterogeneous clients to improve performance~\citep{oh2022fedbabu, dong2022spherefed}. The fixed classifier serves as a consistent criterion for learning representations across clients over time, rather than being an optimal choice itself. For instance, \cite{yang2022inducing} proposes to initialize the classifier as a simplex equiangular tight frame (ETF) with maximal pairwise angles under imbalanced learning. In general, these fixed classifiers force the feature prototypes to converge to an optimal structure to improve representation learning. However, their effectiveness in resolving Fed-LT regarding both global and local models is not investigated.

We examine the effectiveness of training with fixed classifiers in Fed-LT from the perspective of neural collapse (NC)~\citep{papyan2020prevalence}. 
The NC identifies a salient property of the feature space, showing that all within-same-class features tend to collapse to their respective class means. However, as shown in Fig.~\ref{fig:mean_std}(a)\footnote{{Our $\psi_{SSE-C}$ leads to two notable improvements in Fig. 1 (b) over Fig. 1 (a): First, it \textbf{masks noisy features} with large variances, now in sparse areas marked by grey. Second, the role of \textbf{dominant features} is enhanced, as shown by reduced variances among those with larger means, reflecting their increased precision and efficacy.}
}, preliminary experiments benchmarking ETF with CIFAR-100 in Fed-LT suggest that only a few features have relatively large means, while most of the small-mean features are contaminated by severe noise. Such observations are inconsistent with the feature collapse property, and we coin it as \textit{feature degeneration}. More details of the computation process for the data in Fig.~\ref{fig:mean_std} as well as the necessary explanation can be found in Appendix A.2.

To resolve the feature degeneration for improved representation learning, we propose the Static Sparse Equiangular Tight Frame Classifier (SSE-C) inspired by the sparse coding theories~\citep{frankle2018the, glorot2011deep}. The assumption of SSE-C is that the small-mean degenerated features contribute little to model performance while forcing sparsity on them would help learn more expressive representations. We refer to the small-mean features as negligible features and the large-mean features as dominant features.
The SSE-C dynamically prunes the classifier weights of those small-mean noisy features, while holding more expressive dominant features. Probing into weights trained with SSE-C validates our assumption, as shown in Fig.~\ref{fig:mean_std} (b) and experiments.

\noindent \texttt{C2:} \textit{How to conduct effective feature realignment to improve the performance of both the generic and personalized models based on data preferences?}

In Fed-LT, the long-tailed global data distribution and heterogeneous local distributions raise the requirements to learn different global and personalized local models for satisfactory performance. Thus, the feature extractor trained with a fixed classifier needs to be realigned to both global and local models. For the global model, it is necessary to realign the model to improve its performance on global tail classes. For the personalized model, the classifier needs to align features with respective heterogeneous local data preferences.  

\begin{wrapfigure}{r}{0.55\textwidth}
  \vspace{-10pt}
  \centering
    \includegraphics[width=0.55\textwidth]{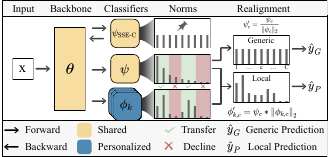}
  \vspace{-10pt}
  \caption{
The framework of \texttt{FedLoGe}. The SSE-C enhances the capability of the backbone (feature extractor); during feature alignment, the model transfers the most crucial information from the global classifier to the personalized model, omitting information pertaining to categories with low usage.}

  \label{fig:pnorm_motivation}
  \vspace{-10pt}
\end{wrapfigure}

In this work, we unify the feature realignment for both the server and clients under the neural collapse framework. The key idea is to align both global and local classifiers based on the weight norm of the classifiers. Previous works show that classifier weight norms are closely correlated with the corresponding class cardinalities \citep{kang2019decoupling, tan2021equalization, li2020overcoming}.
Further research provides both empirical explanation and theoretical justification that the classifier weight norm is larger for majority classes while smaller for minorities \citep{kim2020adjusting,dang2023neural, thrampoulidis2022imbalance}. 

We propose the Global and Local Adaptive Feature realignment (GLA-FR) module to align the backbone trained with SSE-C to the server and clients. In particular, we devise auxiliary classifier heads for the global ($\mathbf{\psi}$ ) and $K$ local classifiers ($\{\mathbf{\phi}_k\}_{k=1}^K$), which are trained alternately with SSE-C in each epoch, see Algorithm \ref{alg:fedloge}.
The alignment includes two stages: global alignment and local alignment. The global realignment is simple yet effective, adjusting the weights based on the norms of $\mathbf{\psi}$ (Eq.\ref{equa:global_realignment}) to tackle with global balanced test set. The alignment for personalized models is a bit different, as their 
data distributions differ greatly from the global distribution. We integrate the global trends with each local client's preference by adjusting the global classifiers $\mathbf{\psi}$ with the norms of local classifier $\mathbf{\phi}_k$ (Eq.\ref{equa:local_realignment}).

Our work represents pioneering efforts to achieve a harmonious integration of global and personalized model learning under Fed-LT, thereby facilitating each participating institution in obtaining a model that is more adeptly tailored to its inherent characteristics and preferences. Comprehensive experiments on representative datasets CIFAR-10/100-LT, ImageNet, and iNaturalist demonstrate the outperformance and efficacy of both global generic and local personalized Fed-LT models.

\section{Related Work}
\subsection{Federated Learning}
\noindent \textbf{Federated Long-tailed Learning} 
Recent research launched attempts to resolve the Fed-LT task. Model decoupling methods explore frameworks such as classifier retraining~\citep{shang2022federated} and prototype-based classifier rebalancing \citep{yang2023integrating, dai2023tackling} for Fed-LT. \cite{shang2022fedic} and \cite{wang2022logit} investigate calibration and distillation methods to improve the model performance. Many efforts also have been made from the perspective of meta-learning \citep{qian2023long, shen2021agnostic}, client selection \citep{zhang2023fed, yang2021federated}, re-weighting \citep{wang2021addressing, shen2021agnostic}, and aggregation~\citep{chou2022grp}.

\noindent \textbf{Personalized Federated Learning (pFL)}
To deal with the poor generalization performance of the single generic global at local data, pFL has been vastly investigated.
A group of works seeks to train local personalized models via transferring knowledge from the generic global model~\citep{li2019fedmd, t2020personalized, fallah2020personalized,chen2023spectral}. 
Multi-task learning-based methods are also been explored with client clustering \citep{sattler2020clustered, briggs2020federated, ghosh2020efficient} and model interpolation \citep{deng2020adaptive, li2021ditto, diao2020heterofl}.

For neural network-based FL framework, parameter decoupling methods have gained popularity due to their simplicity. Parameter decoupling aims to achieve personalization by decoupling the local private model parameters from the global model parameters. For horizontal decoupling, \cite{li2021fedbn} personalizes the batch normalization layers, \cite{pillutla2022federated} explores decoupling different parts, and personalizing the last layer is adopted in \cite{arivazhagan2019federated}, \cite{collins2021exploiting}, and \cite{briggs2020federated}. For vertical decoupling, \cite{shen2022cd2} personalizes channels.

\subsection{Neural Collapse for Representation Learning}

Neural collapse refers to a set of four interconnected phenomena that demonstrate a pervasive inductive bias in the terminal phase of training, as shown by \cite{papyan2020prevalence}. Subsequently, several works have sought to explain the neural collapse phenomena from the perspective of peeled models \citep{ji2021unconstrained, fang2021exploring}, unconstrained feature models \citep{tirer2022extended, mixon2020neural, zhu2021geometric}, and Riemannian manifolds \citep{yaras2022neural}. 
Building upon the findings on neural collapse, \cite{yang2022inducing} first proposed fixing the classifier to an ETF structure and introduced dot regression loss. The ETF structure was later utilized for semantic segmentation \citep{zhong2023understanding}, handling heterogeneity in federated learning \citep{li2023no}, transfer learning \citep{li2022principled}, incremental learning \citep{yang2023neural}, and object detection \citep{ma2023digeo}.


In summary, our method stands out by introducing personalization within the neural collapse framework, effectively overcoming global imbalanced data and enhancing local model personalization through tailored feature distribution alignment. Contrary to FedETF's fixed classifier approach and FedRod's dual classifier structure, our method ensures superior handling of global long-tail bias and precise local model tuning, marking a novel advancement in Fed-LT. For detailed comparisons, please see Section~\ref{subsec:comparison} in the Appendix.

\section{Proposed method}
In this section, we introduce our proposed \texttt{Fed-LoGe} (see Algorithm~\ref{alg:fedloge}), a simple yet effective framework to achieve joint personalized and generic model learning for Fed-LT. To boost representation learning and address the global-local inconsistency, we introduce a training paradigm consisting of a sparsified ETF module and global-local feature alignment modules.

\subsection{Preliminaries}
We consider an FL system with $K$ clients and a server. The overall objective is to train $1+K$ models: $1$ generic and $K$ personalized models. Specifically, the generic model is parameterized by $w=\{\theta,\mathbf{\psi}\}$, whereas the $k^{\text{th}}$ personalized model for each client $k\in[K]$ is denoted as $w_k$.  
We decoupled the neural network models into a feature extractor $f(x, \theta)$ and a set of classifiers. The feature extractor, parameterized by $\theta$, transforms input $x$ into features $h$. The generic classifier $g(h, \mathbf{\psi})$ and personalized classifiers $g(h, \mathbf{\phi}_k)$ then map these features to the output labels. 
The overall global and local objective functions can be respectively expressed as:
\begin{equation}\label{equa:obj_leader}
\textbf{Global: }\min_{w} \sum_{k=1}^{K} \frac{|\mathcal{D}_k|}{|\mathcal{D}|} \mathcal{L}_k(w_k, \mathcal{D}_k), \quad \textbf{Local: } \min_{\{\theta, \mathbf{\phi}_k\}} \mathcal{L}(\theta, \mathbf{\phi}_k; \mathcal{D}_k), 
\end{equation}
where $\mathcal{D}=\{\mathcal{D}_k\}_{k=1}^K$ is the global long-tailed dataset composed of $K$ heterogeneous local datasets, each with size $|\mathcal{D}_k|$. 
The training is executed over $T$ rounds. In each round $t$, the server distributes the current global model $w^{(t)}$ to all clients for local updates. Furthermore, given that $c$ represents the class index, for $\forall c \in C$, the classifier vector is $\mathbf{\psi}_c$, and the corresponding features are $h_c$.

\subsection{Static Sparse Equiangular Tight Frame Classifier (SSE-C)}\label{subsec:sse-c}

Initializing the classifier as ETF and subsequently freezing it during training has proven to be an effective strategy in federated learning, attributed to ETF's inherent structure, which ideally exhibits commendable properties of feature collapse under balanced data. However, we found that in the context of Fed-LT, the feature collapse property was not satisfied when initializing the fixed classifiers due to feature degeneracy (the existence of a high-noise feature with a small norm), which is illustrated in Fig.~\ref{fig:mean_std} (a). 
Accordingly, we propose the Static Sparse Equiangular Tight Frame Classifier (SSE-C) via fixing classifier to learn higher-quality features and reduce the impact of negligible features so as to achieve effective representation learning. The server will obtain SSE-C with $\mathcal{L}_{\text{SSE-C}}$ prior to the local training, and all clients will fix SSE-C throughout the training. 

We first initialize the classifier as a conventional ETF matrix by 

\vspace{-10pt}
\begin{equation}\label{equa:etf_constru}
    \mathbf{\psi}=\sqrt{\frac C{C-1}}\mathbf{U}\left(\mathbf{I}_C-\frac1C\mathbf{1}_C\mathbf{1}_C^T\right)
\end{equation}
\vspace{-10pt}

where $\mathbf{\psi}=[\mathbf{\psi}_{:,1},\cdots,\mathbf{\psi}_{:,C}]\in\mathbb{R}^{d\times C},\mathbf{U}\in\mathbb{R}^{d\times C} $ allows any rotation and satisfies $\mathbf{U}^T\mathbf{U}=\mathbf{I}_C,\mathbf{I}_C$ is the identity matrix. $d$ is the dimension of the classifier vector, and $\mathbf{1}_C$ is an all-ones vector. We can deduce the important property that \textit{all class vector has the equal $\ell_{2}$ norm and maximal pair-wise angle $-\frac{1}{C-1}$ in $\mathbb{R}^{d}$}. 
Note that randomly assigning $\beta$ proportion of weights in the ETF matrix to $0$ will disrupt the ETF condition: the class vector angles cease to be maximal and equal, and the norms of the classifier vector become unequal. As such, it is necessary to train a sparse ETF structure that satisfies the ETF geometric conditions. 
We introduce a sparse indicator matrix $\mathbf{S}$ with the same dimensions as $\mathbf{\psi}$, where $\beta$ proportion of the elements are randomly set to $0$. Then, the sparsified matrix $\mathbf{\psi}^{\prime}$ can be represented as $\mathbf{\psi}^{\prime}=\mathbf{\psi}\odot\mathbf{S}$.
We design the \textit{Equal Norm Loss} and \textit{Maximal Angle Loss} to optimize the geometric structure to meet the conditions of ETF. First, the \( \ell_2 \) norm of class vectors should be equal. We constrain all norms of class vector values to a predetermined $\gamma$: 

\vspace{-10pt}
\begin{equation}\label{equa:loss_norm}
    l_{\mathrm{norm}}(\mathbf{\psi}^{\prime},\gamma, \mathbf{S})=\sum_{i=1}^C\left(\left\|\mathbf{\psi}^{\prime}_{:,i}\odot\mathbf{S}_{:,i}\right\|_2-\gamma\right)^2.
\end{equation}
\vspace{-10pt}

Second, we maximize the minimum angle between class vector pairs. 
Following MMA \citep{wang2020mma}, we normalize the classifier vector by $\hat{\mathbf{\psi}}^{\prime}_{:,i}=\frac{\mathbf{\psi}^{\prime}_{:,i}}{\left\|\mathbf{\psi}^{\prime}_{:,i}\right\|_2}$ and \textit{maximize only the minimum angle} with the formula:

\vspace{-10pt}
\begin{equation}\label{equa:loss_angle}
    l_\text{angle}(\mathbf{\hat{\mathbf{\psi}}^{\prime}},\mathbf{S})=-\frac1C\sum_{i=1}^C\cos^{-1}\left(\max_{j\in\{1,2,\ldots,C\}\setminus\{i\}}\left((\mathbf{\hat{\mathbf{\psi}}^{\prime}}_{:,i}\odot\mathbf{S}_{:,i})^T(\mathbf{\hat{\mathbf{\psi}}^{\prime}}_{:,j}\odot\mathbf{S}_{:,j})\right)\right).
\end{equation}
\vspace{-10pt}

By integrating the $l_{\text{norm}}$ and $l_{\text{angle}}$ , we obtain the $ \mathcal{L}_{\text{SSE-C}}$ for the following training:

\vspace{-10pt}
\begin{equation}\label{equa:l_etf}
    \mathcal{L}_{\text{SSE-C}} = l_{\text{norm}}(\mathbf{\psi}^{\prime},\gamma,\mathbf{S}) + l_{\text{angle}}(\hat{\mathbf{\psi}^{\prime}},\mathbf{S}).
\end{equation}
\vspace{-10pt}

Then we solve the objective $\mathbf{\psi}_\text{SSE-C} = \arg\min_\mathbf{\psi}{\mathcal{L}_{\text{SSE-C}}}$ by SGD. By regularizing class vector norms and maximizing their minimum angle, the classifier exhibits sparsity while maintaining ETF properties, effectively guiding the model in learning robust features.


\begin{algorithm}[t!]
    \caption{An overview of \texttt{FedLoGe} framework}
    \label{alg:fedloge}
    \renewcommand{\algorithmicrequire}{\textbf{Input:}}
    \renewcommand{\algorithmicensure}{\textbf{Output:}}
    \algrenewcommand\algorithmicreq{\textbf{function}}
    
    \begin{algorithmic}[1]
        \REQUIRE $w^{\{0\}} = \{\theta^{(0)}, \mathbf{\psi}^{(0)}\}$,$\{\mathbf{\phi}_k^{(0)}\}_{k=1}^K$, $K, E, T, C$ 
        \ENSURE $w^{\{T\}} = \{\theta^{(T)}, \mathbf{\psi}^{\prime (T)}\}$, $\{w_k\}_{k=1}^K=\{\theta^{(T)},\mathbf{\phi}_k^{\prime (T)}\}_{k=1}^K$
        \STATEx \tikz[baseline=(current bounding box.center)] {
    \node[fill=mypink, draw=mydarkpink, line width=0.3pt, rounded corners=2pt, inner sep=1pt] 
    {\textbf{Stage 1: Representation Learning with SSE-C}};
}
        \STATE Obtain $\mathbf{\psi}_\text{SSE-C}$ by Equation \ref{equa:l_etf}
        \FOR{each round $t=1$ \textbf{to} $T$}
            \STATE $S_t \leftarrow $ subset of selected clients
            \FOR{each client $k$ $\in$ $S_t$ \textbf{in parallel}}
                \STATE $\theta^{t}_k, \mathbf{\psi}^{t}_k$ $\leftarrow$ \textsc{ClientUpdate}($k, \theta^{(t-1)}, \mathbf{\psi}^{(t-1)}, \mathbf{\psi}_\text{SSE-C}$)
            \ENDFOR
            \STATE $\theta^{(t)}, \mathbf{\psi}^{(t)} = \textsc{Aggregation}(\theta^{(t-1)}_k, \mathbf{\psi}^{(t-1)}_k)$, for all $k \in S_t$
        \ENDFOR

    \STATEx \tikz[baseline=(current bounding box.center)] {
    \node[fill=mypink, draw=mydarkpink, line width=0.3pt, rounded corners=2pt, inner sep=1pt] 
    {\textbf{Stage 2: Global Feature Realignment}};
}
    \FOR{each classifier vector $c=1$ \textbf{to} $C$}
        \STATE $\mathbf{\psi}^{\prime (T)}_c \leftarrow \mathbf{\psi}^{(T)}_c/\left\| \mathbf{\psi}^{(T)}_{c} \right\|_2$ (Equation \ref{equa:global_realignment})
        \hspace{0.4cm} \textit{$\backslash\backslash$ Align the long-tailed norm to balanced norm}
    \ENDFOR

    \STATEx \tikz[baseline=(current bounding box.center)] {
    \node[fill=mypink, draw=mydarkpink, line width=0.3pt, rounded corners=2pt, inner sep=1pt] 
    {\textbf{Stage 3: Local Feature Realignment}};
}
    \FOR{each client $k=1$ \textbf{to} $K$}
        \FOR{each classifier vector $c=1$ \textbf{to} $C$}
            \STATE $\mathbf{\phi}^{\prime (T)}_{k, c} \leftarrow \mathbf{\psi}^{(T)}_{c} * \left\| \mathbf{\phi}^{(T)}_{k, c} \right\|_2$ (Equation \ref{equa:local_realignment}) \hspace{0cm} \textit{$\backslash\backslash$ Incorporate Global Classifier with local statistics}
        \ENDFOR

        \STATE $\mathbf{\phi}^{\prime (T)}_k \leftarrow \mathbf{\phi}^{\prime (T)}_k - \eta \nabla_{\mathbf{\phi}}(\mathcal{L}(\theta^{T}, \mathbf{\phi}^{\prime (T)}_k); x)$ \hspace{0.5cm} \textit{$\backslash\backslash$ Finetune $\mathbf{\phi}^{\prime (T)}_k$}
    \ENDFOR
    \RETURN $w^{(T)} = \{\theta^{(T)}, \mathbf{\psi}^{(T)}\}$, $\{w_k\}_{k=1}^K=\{\theta^{(T)},\mathbf{\phi}_k^{(T)}\}_{k=1}^K$
    \end{algorithmic}

        \begin{algorithmic}[1]
    \REQ{\textsc{ClientUpdate}($k, \theta^{(t)}, \mathbf{\psi}^{(t)}, \mathbf{\psi}_\text{SSE-C}$)}
    \STATE $\theta^{(t)}_k, \mathbf{\psi}^{(t)}_k = \theta^{(t)}, \mathbf{\psi}^{(t)} $ 
    \FOR{each local epoch $i=1$ \textbf{to} $E$}

        \STATE Compute features $h_i \leftarrow f(x_i, \theta^{(t)})$

        \STATE $\theta^{(t+1)}_k \leftarrow \theta^{(t)}_k - \eta \nabla_{\theta}(\mathcal{L}(\mathbf{\psi}_{\text{SSE-C}}; x_i))$ \hspace{0.87cm} \textit{\( \backslash\backslash \) Fix \( \mathbf{\psi}^{(t)}_k \), \( \mathbf{\phi}^{(t)}_k \), Update \( \theta^{(t)}_k \) with \( \mathbf{\psi}_\text{SSE-C} \)}

        \State \hspace*{-0.35cm} $\left.\begin{array}{l}
            \mathbf{\psi}^{(t+1)}_k \leftarrow \mathbf{\psi}^{(t)}_k - \eta \nabla_{\mathbf{\psi}}(\mathcal{L}(\theta^{t}, \mathbf{\psi}^{(t)}_k); h_i)\\
            \mathbf{\phi}^{(t+1)}_k \leftarrow \mathbf{\phi}^{(t)}_k - \eta \nabla_{\mathbf{\phi}}(\mathcal{L}(\theta^{t}, \mathbf{\phi}^{(t)}_k); h_i)
        \end{array}\right\} \hspace{0.35cm} \textit{\( \backslash\backslash \) Fix \( \theta^{(t)}_k \), Update \( \mathbf{\psi}^{(t)}_k \), \( \mathbf{\phi}^{(t)}_k \)}$


    \ENDFOR
    \RETURN $\theta^{(t+1)}_k$, $\mathbf{\psi}^{(t+1)}_k$
            
   \end{algorithmic}
    
\end{algorithm}

\subsection{Global and Local Adaptive Feature Realignment (GLA-FR)}\label{subsec:gla-fr}

To be adapted in both global and local models, the feature extractor trained with a fixed classifier needs to be realigned to address the imbalance and heterogeneity.
Hence, we conduct feature realignment to both global and personalized models after training the SSE-C guided feature extractor, where the realignment should be consistent with the local data statistics/class cardinality.

To obtain a good estimation of class cardinality, in prior work such as \cite{kang2019decoupling, tan2021equalization, li2020overcoming}, the classifier weight norms $\lVert \mathbf{\psi}_c \rVert$ are found to be correlated with the corresponding class cardinalities $n_c$, in which $\mathbf{\psi}_c$ is the classifier weight vector for the $c$-th class. \cite{kim2020adjusting} provides an explanation from the perspective of decision boundaries - the weight vector norm for more frequent classes is larger, biasing the decision boundary towards less frequent classes. Also, for the neural collapse framework with imbalanced data, the relations between the weight norm of classifiers and the class cardinality also exist \citep{thrampoulidis2022imbalance,dang2023neural}. These findings are consistent, which motivates us to measure/estimate local data statistics based on the norm weight of the classifier.

The frozen ETF classifier is not suitable for feature alignment, attributed to the lack of valid norms to estimate class cardinality. We design a new auxiliary global head $\mathbf{\psi}$ to obtain valid norms, which participate in gradient updates and weight aggregation alongside the backbone. 
After $T$ rounds of training, we get the global weight $w^{(T)}=\{\theta^{(T)},\mathbf{\psi}^{(T)}\}$. The $\theta^{(T)}$ is well trained with $\mathbf{\psi}_{SSE-C}$.


For the global adaptive feature distribution process (GA-FR), let $\mathbf{\psi}_{c}$ denote a classifier vector corresponding to the $c^{\text{th}}$ class, where $ c \in C$. The aligned classifier vector \( \mathbf{\psi}_{c}^\prime \) can be obtained by dividing \( \mathbf{\psi}_{c} \) by its \( \ell_2 \) norm as follows:
\begin{equation}\label{equa:global_realignment}
    \mathbf{\psi}_{c}^\prime = \mathbf{\psi}_{c}/\left\| \mathbf{\psi}_{c} \right\|_2
\end{equation}
\vspace{-10pt}

Here, \( \left\| \mathbf{\psi}_{c} \right\|_2 \) represents the \( \ell_2 \) norm of \( \mathbf{\psi}_{c} \). Each \( \mathbf{\psi}_{c}^\prime \) will be a unit vector, preserving the direction of \( \mathbf{\psi}_{c} \) and possessing the magnitude of $1$.

For personalized adaptive feature realignment (LA-FR), we adapt the global auxiliary classifier $\mathbf{\psi}$ to the personalized classifier by multiplying the norm of $\mathbf{\phi}_k$, which implies that clients will leverage information from categories with a larger sample size while omitting information pertaining to the rare categories.
For local classifier vector $\mathbf{\phi}_{k,c}$ at client $k$, the process of LA-FR is as follows: 

\begin{equation}\label{equa:local_realignment}
    \mathbf{\phi}^{\prime}_{k,c}=\mathbf{\psi}_{c} * \left\| \mathbf{\phi}_{k,c} \right\|_2
\end{equation}
\vspace{-15pt}

\subsection{Algorithms}
Overall, our framework \texttt{Fed-LoGe} consists of three critical stages: representation learning with SSE-C, global feature realignment, and local feature realignment, for the training of the shared backbone $\theta$, global auxiliary classifier $\mathbf{\psi}$, and $K$ local classifiers ($\{\mathbf{\phi}_k\}_{k=1}^K$), respectively.

In the first stage, the server first constructs the SSE-C with Eq.~\ref{equa:l_etf} before the training and then distributes it to all clients. Upon receiving SSE-C, each client fixes it as the classifier to train the backbone \( \theta \), global classifier $\mathbf{\psi}$, and local classifier $\mathbf{\phi}_k$ alternately. Specifically, we update $\theta$ with fixed $\mathbf{\psi}_\text{SSE-C}$. Subsequently, the $\theta$ is frozen to update the global head \( \mathbf{\psi} \) and each local classifier \( \mathbf{\phi}_k \). At the end of each round, the \( \theta \) and \( \mathbf{\psi} \) are aggregated at the server, while \( \mathbf{\phi}_k \) is retained locally. 

Global adaptive feature realignment (GA-FR) is performed in the second stage, where each class vector is redistributed by the server according to its individual norms, as outlined by Eq.~\ref{equa:global_realignment}. Subsequently, in the third phase, personalized adaptive feature realignment (LA-FR) for the class vectors of the global auxiliary head $\mathbf{\psi}$ is performed. Following LA-FR, local finetuning could be further conducted to boost the model performance. A summary of \texttt{Fed-LoGe} is given in Algorithm \ref{alg:fedloge}.
 
\renewcommand\arraystretch{1}
\begin{table*}[t!]
\centering
\begin{adjustbox}{width=0.95\textwidth,center}
\begin{tabular}{llcccccccc}
\hline
\multirow{3}{*}{Dataset}       & Non-IID          & \multicolumn{4}{c}{$\alpha=0.5$}                                 & \multicolumn{4}{c}{$\alpha=1$}                                   \\ \cline{2-10} 
                               & Imbalance Factor & \multicolumn{2}{c}{IF=50}     & \multicolumn{2}{c}{IF=100}    & \multicolumn{2}{c}{IF=50}     & \multicolumn{2}{c}{IF=100}    \\ \cline{2-10} 
                               & Method/Model     & GM            & PM            & GM            & PM            & GM            & PM            & GM            & PM            \\ \hline
\multirow{11}{*}{CIFAR-10-LT}  & FedAvg           & 0.7988        & 0.8722        & 0.7214        & 0.8664        & 0.7890        & 0.8916        & 0.7231        & 0.8935        \\
                               & FedProx          & 0.7869        & 0.8653        & 0.7160        & 0.8617        & 0.7790        & 0.8888        & 0.7266        & 0.8897        \\
                               & FedBN            & 0.7604        & 0.8837        & 0.6984        & 0.8847        & 0.7596        & 0.8916        & 0.7033        & 0.8925        \\
                               & FedPer           & -             & 0.8935        & -             & 0.8931        & -             & 0.8918        & -             & 0.8999        \\
                               & FedRep           & 0.7938        & 0.8988        & 0.7218        & 0.8984        & 0.7816        & 0.9043        & 0.7271        & 0.9065        \\
                               & Ditto            & 0.7813        & 0.8926        & 0.7180        & 0.8874        & 0.7804        & 0.8967        & 0.7210        & 0.8973        \\
                               \cdashline{2-10}[4pt/3pt]
                               & FedROD           & 0.7862        & 0.9030        & 0.7137        & 0.8941        & 0.7776        & 0.9025        & 0.7236        & 0.9041        \\
                               & FedBABU          & 0.7851        & 0.8664        & 0.7280        & 0.8613        & 0.7819        & 0.8914        & 0.7316        & 0.8898        \\
                               & FedETF           & 0.8056        & 0.7446        & 0.6709        & 0.7323        & 0.7453        & 0.8106        & 0.6615        & 0.7917        \\
                               & Ratio Loss       & 0.7995        & 0.8791        & 0.7290        & 0.8724        & 0.7857        & 0.8938        & 0.7342        & 0.8934        \\ \cline{2-10} 
                               & \cellcolor{pink!40} FedLoGe             & \cellcolor{pink!40}$\bm{0.8277}$ & \cellcolor{pink!40}$\bm{0.9112}$ & \cellcolor{pink!40}$\bm{0.7672}$ & \cellcolor{pink!40}$\bm{0.9096}$ & \cellcolor{pink!40}$\bm{0.8189}$ & \cellcolor{pink!40}$\bm{0.9104}$ & \cellcolor{pink!40}$\bm{0.7593}$ & \cellcolor{pink!40}$\bm{0.9073}$ \\ \hline \hline
\multirow{11}{*}{CIFAR-100-LT} & FedAvg           & 0.4271        & 0.6019        & 0.3818        & 0.6214        & 0.4215        & 0.6086        & 0.3797        & 0.6269        \\
                               & FedProx          & 0.4276        & 0.6071        & 0.3856        & 0.6214        & 0.4187        & 0.6089        & 0.3788        & 0.6261        \\
                               & FedBN            & 0.4332        & 0.6657        & 0.3895        & 0.6703        & 0.4229        & 0.6373        & 0.3847        & 0.6562        \\
                               & FedPer           & -             & 0.6888        & -             & 0.7007        & -             & 0.6476        & -             & 0.6737        \\
                               & FedRep           & 0.4355        & 0.6911        & 0.3907        & 0.6917        & 0.4283        & 0.6539        & 0.3828        & 0.6831        \\
                               & Ditto            & 0.4312        & 0.6441        & 0.3847        & 0.6588        & 0.4188        & 0.6182        & 0.3822        & 0.6320        \\
                               \cdashline{2-10}[4pt/3pt]
                               & FedROD           & 0.4384        & 0.7124        & 0.3919        & 0.6919        & 0.4289        & 0.6667        & 0.3902        & 0.6917        \\
                               & FedBABU          & 0.4415        & 0.6416        & 0.3921        & 0.6480        & 0.4336        & 0.6354        & 0.3907        & 0.6598        \\
                               & FedETF           & 0.4223        & 0.6055        & 0.3825        & 0.6421        & 0.4278        & 0.6302        & 0.4278        & 0.6507        \\
                               & Ratio Loss       & 0.4326        & 0.6152        & 0.3912        & 0.6348        & 0.4253        & 0.6213        & 0.3839        & 0.6344        \\ \cline{2-10} 
                               & \cellcolor{pink!40} FedLoGe             & \cellcolor{pink!40}$\bm{0.4762}$ & \cellcolor{pink!40}$\bm{0.7229}$ & \cellcolor{pink!40}$\bm{0.4233}$ & \cellcolor{pink!40}$\bm{0.7285}$ & \cellcolor{pink!40}$\bm{0.4860}$ & \cellcolor{pink!40}$\bm{0.7099}$ & \cellcolor{pink!40}$\bm{0.4330}$ & \cellcolor{pink!40}$\bm{0.7195}$ \\ \hline
\end{tabular}
\end{adjustbox}
\caption{Test accuracies of our and SOTA methods on CIFAR-10/100-LT with diverse imbalanced and heterogeneous data settings. GM/PM denotes Global/Personalized model.}
\vspace{-10pt}
\label{tab:cifar10_100_acc}
\end{table*}

\section{Experiments}
\subsection{Experimental Setup}

\noindent\textbf{Dataset, Models and Metrics:}
We consider image classification tasks for performance evaluation on benchmark long-tailed datasets: CIFAR-10/100-LT, ImageNet-LT, and iNaturalist-User-160k \citep{van2018inaturalist}. The CIFAR-10/100-LT datasets are sampled into a long-tailed distribution employing an exponential distribution governed by the Imbalance Factor ($\text{IF}$) in \cite{cao2019learning}.  
All experiments are conducted with non-IID data partitions, implemented by the Dirichlet distributions-based approach with parameter $\alpha$ to control the non-IIDness \citep{chen2022on}.
ResNet-18 is trained over $K=40$ clients on CIFAR-10-LT, while ResNet-34 and ResNet-50 are implemented on CIFAR-100-LT and ImageNet-LT, respectively, with $K=20$ clients. The configurations for iNaturalist-160k align with those utilized for ImageNet-LT.
We use $\alpha=1, 0.5$ and $\text{IF}$ = $50, 100$ in CIFAR-10/100-LT.  We use $\alpha= 0.1, 0.5$ in ImageNet-LT and iNaturalist, respectively.

A global balanced dataset is used for the calculation of test accuracy to evaluate global model (GM) performance. We also report the accuracy across many, medium, and few classes. The detailed categorization for many/med/few classes can be found in the Appendix.

For personalized model (PM) evaluation, we use local test accuracy, and the local test set is sampled from the global test set.
Each local test set has an identical distribution to the local training set. The accuracy of the PM is the arithmetic mean of local test accuracy across all clients.

\renewcommand\arraystretch{1.2}
\begin{table*}[t!]
\centering
\begin{adjustbox}{width=0.85\textwidth,center}
\begin{tabular}{lcccccccccc}
\hline
Dataset      & \multicolumn{5}{c}{ImageNet}                    & \multicolumn{5}{c}{Inaturalist}                                   \\ \hline
Method/Model & Many   & Med    & Few    & GM     & PM           & Many  & Med          & Few          & GM           & PM           \\ \hline
FedAvg       & 0.481  & 0.307  & 0.159  & 0.329  & 0.528        & 0.591 & 0.418        & 0.238        & 0.425        & 0.590        \\
FedProx      & 0.493  & 0.318  & 0.180  & 0.343  & 0.500        & 0.525 & 0.484        & 0.223        & 0.432        & 0.596        \\
FedBN        & 0.471  & 0.300  & 0.168  & 0.319  & 0.504        & 0.573 & 0.396        & 0.221        & 0.413        & 0.563        \\
FedPer       & -      & -      & -      &        & 0.653        & -     & -            & -            & -            & 0.638        \\
FedRep       & 0.460  & 0.309  & 0.187  & 0.330  & 0.574        & 0.571 & 0.453        & 0.237        & 0.429        & 0.627        \\ \cdashline{1-11}[4pt/3pt]
Ditto        & 0.492  & 0.319  & 0.176  & 0.342  & 0.674        & $\bm{0.598}$ & 0.452        & 0.243        & 0.437        & 0.584        \\
FedROD       & 0.483  & 0.305  & 0.165  & 0.331  & 0.7033       & 0.585 & 0.416        & 0.243        & 0.421        & 0.699        \\
FedBABU      & 0.443  & 0.240  & 0.055  & 0.230  & 0.425        & 0.561 & 0.401        & 0.199        & 0.377        & 0.696        \\
FedETF       & 0.425  & 0.239  & 0.05   & 0.222  & 0.418        & 0.587 & 0.431        & 0.245        & 0.437        & 0.713        \\
Ratio Loss   & $\bm{0.495}$  & 0.337  & 0.189  & 0.351  & 0.521        & 0.587 & 0.454        & 0.290        & 0.452        & 0.589        \\ \hline 
\rowcolor{pink!40}
FedLoGe         & 0.430 & $\bm{0.373}$ & $\bm{0.285}$ & $\bm{0.356}$ & $\bm{0.726}$ & 0.519 & $\bm{0.508}$ & $\bm{0.473}$ & $\bm{0.503}$ & $\bm{0.759}$ \\ \hline
\end{tabular}
\end{adjustbox}
\caption{Test accuracies of our and SOTA methods on ImageNet-LT and iNaturalist-160k with diverse heterogeneous data settings. }
\vspace{-10pt}
\label{tab:imagenet_inat_acc}
\end{table*}

\begin{wraptable}{r}{0.45\textwidth}
\vspace{-8pt}
\centering
\resizebox{\linewidth}{!}{
\begin{tabular}{cccccc}
\hline
ETF          & SSE-C        & GA-FR        & LA-FR        & GM & PM \\ \hline
$\checkmark$ &              &              &              & 0.3825   &  0.6421  \\
             & $\checkmark$ &              &              & 0.4175   &  0.6865  \\
             & $\checkmark$ & $\checkmark$ &              & $\bm{0.4350}$   &  0.6868  \\
             & $\checkmark$ &              & $\checkmark$ & 0.4175   & 0.7339   \\
$\checkmark$ &              & $\checkmark$ & $\checkmark$ &  0.3988   & 0.7043   \\
             & $\checkmark$ & $\checkmark$ & $\checkmark$ & $\bm{0.4350}$  & $\bm{0.7343}$   \\ \hline
\end{tabular}
}
\caption{Ablations of SSE-C and GLA-FR.}
\label{tab:ablations}
\vspace{-15pt}
\end{wraptable}

\noindent\textbf{Compared Methods:}
In addition to \texttt{FedAvg} and \texttt{FedProx}~\citep{li2020federated} which are included for reference, we consider two types of state-of-the-art baselines:
(1) pFL methods, including \texttt{FedBN}~\citep{li2021fedbn}, \texttt{FedPer}~\citep{arivazhagan2019federated}, \texttt{FedRep}~\citep{collins2021exploiting}, \texttt{Ditto}~\citep{li2021ditto}, and \texttt{FedROD}~\citep{chen2022on}.
(2) Federated (Long-tailed) Representation learning, including \texttt{FedBABU}~\citep{oh2022fedbabu}, \texttt{FedETF}~\citep{li2023no} and \texttt{Ratio Loss}~\citep{wang2021addressing}.

\subsection{Performance Comparison}
For evaluation on CIFAR-10-LT and CIFAR-100-LT, \texttt{FedLoGe} consistently outperforms all baselines over all settings, achieving the highest overall accuracies in all settings for both GM and PMs; see Tab.~\ref{tab:cifar10_100_acc}.
For ImageNet-LT and iNaturalist-160k, Tab.~\ref{tab:imagenet_inat_acc} highlights that \texttt{FedLoGe} consistently surpasses all baselines, marking significant accuracy improvements, particularly in middle and tail classes.  Overall, benefiting from the enhanced representation learning and classifier realignments, \texttt{FedLoGe} attains superior GM performance, attributed to the SSE-C design for representation learning, and simultaneously obtains impressive PM performance together with refined feature alignment.

\subsection{Ablation study and Sensitivity analysis}
\noindent\textbf{Ablations of SSE-C and GLA-FR:}
In the ablation study, we evaluate the SSE-C and GA-FR/LA-FR individual impacts with respect to both GM and PM performance, over CIFAR-100-LT (IF=100, $\alpha=0.5$), as given in Tab.~\ref{tab:ablations}. The results lead to the following conclusions: Compared to dense ETF, SSE-C can train a superior backbone, enhancing both GM and PM performance. GA-FR can be combined with any backbone to boost GM performance, while LA-FR can be used with any backbone to boost PM performance. Employing SSE-C and GLA-FR simultaneously yields significant enhancements in both GM and PM performance.

\begin{wrapfigure}{r}{0.55\textwidth}
    \centering
    \begin{tabular}{cc}
        \includegraphics[width=0.25\textwidth]{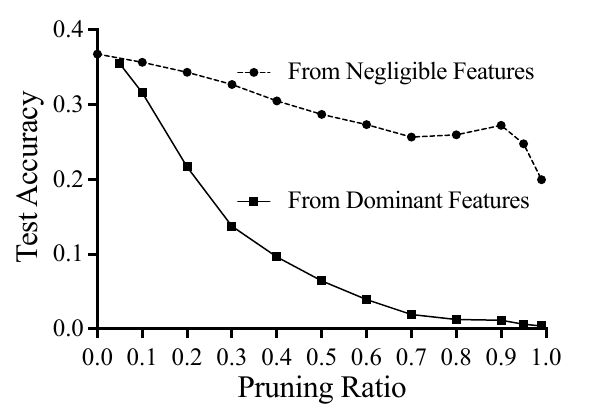} &
        \includegraphics[width=0.25\textwidth]{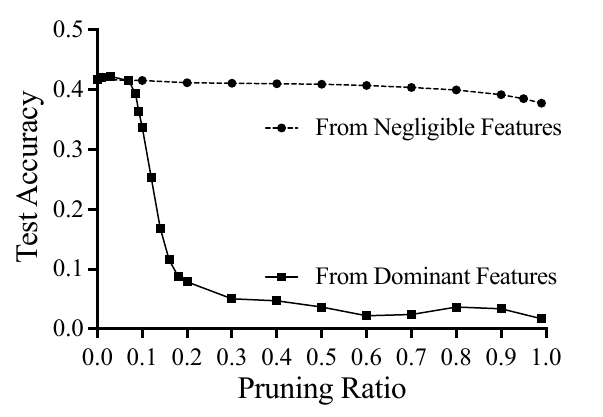} \\
        (a) & (b)
    \end{tabular}
    \caption{(a): Pruning features from smaller means and larger means, respectively. Negligible features exert a minor impact on model performance; (b): Pruning Experiments with SSE-C. The model optimally enhances the dominant features, rendering the impact of negligible features imperceptible.}
    \label{fig:pruning_experiments}
\end{wrapfigure}


\noindent\textbf{Negligible and dominant features with SSE-C:}
We investigated the effects of pruning on different features over CIFAR-100-LT ($\alpha$=0.5, IF=100).
We computed the class mean for each category. Within each class mean, we ordered the means and pruned the corresponding classifier vector weights in descending (from dominant features) and ascending order (from negligible features) with pruning ratios from 0 to $100\%$.
After training with SSE-C, pruning dominant features drastically decreases performance, while pruning negligible features barely affects performance, which is visualized in Fig.~\ref{fig:pruning_experiments} (a, b). This observation indicates that SSE-C can adeptly learn more effective features while autonomously disregarding high-noise features.

\begin{figure}[htbp]
    \centering
    \scriptsize
    \begin{tabular}{cccc}
        \includegraphics[width=0.28\columnwidth]{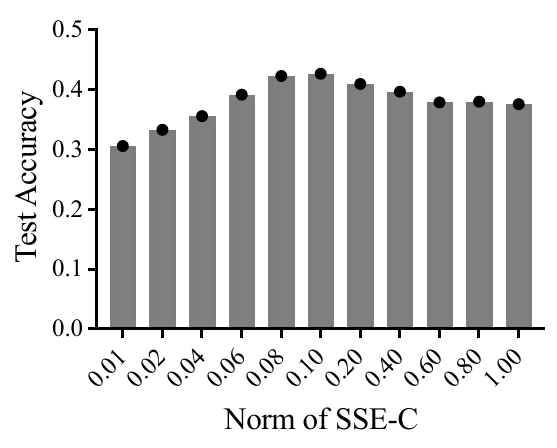}  &
        \includegraphics[width=0.28\columnwidth]{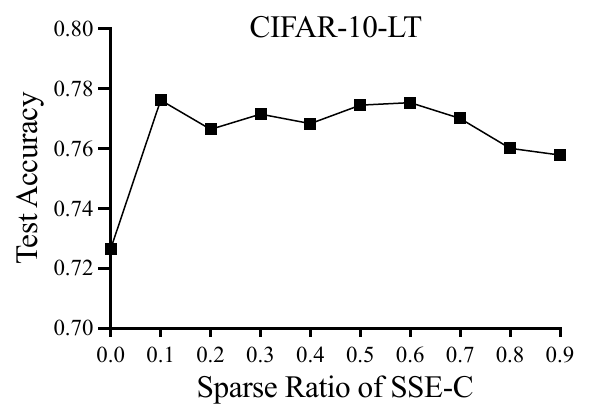}   &
        \includegraphics[width=0.28\columnwidth]{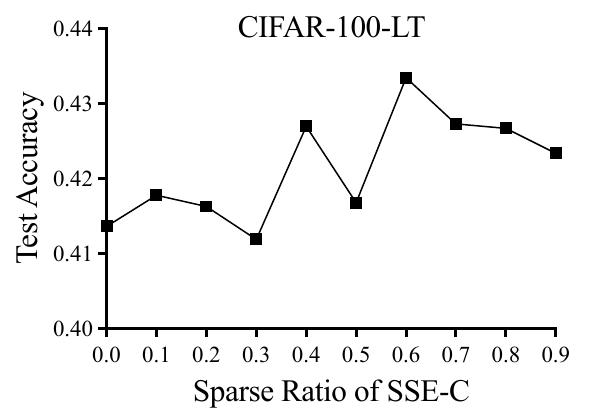}           \\
        \\
        (a) & (b) & (c) 
    \end{tabular}
    \caption{(a): The impact of varying norms $\gamma$ in SSE-C on model performance within CIFAR-100-LT; (b) The impact of sparse ratio $\beta$ on model performance in CIFAR-10-LT; (c): The impact of sparse ratio $\beta$ on model performance in CIFAR-100-LT.}
    \vspace{-5pt}
    \label{fig:norm_sparse_visual_ablation}
\end{figure}

\noindent\textbf{Sensitivity analysis for $\gamma$ in SSE-C:}
We evaluate the model performance with various values of $\gamma$ (which is assigned as the norm of SSE-C in \textbf{norm equal loss}) over CIFAR-100-LT (IF=100, $\alpha=0.5$), as depicted in Fig.~\ref{fig:norm_sparse_visual_ablation} (a). Our observations indicate that a smaller norm leads to a reduced gradient during backpropagation, producing subpar performance. Conversely, a larger norm enables faster convergence. For CIFAR-10/100-LT, the optimal value is approximately 1.0, whereas, for ImageNet and iNaturalist, it is 1.6.

\noindent\textbf{Sensitivity analysis for Sparse Ratio $\beta$ in SSE-C:}
The $\beta$ indicates the pruning proportion in SSE-C. We evaluate the performance with the sparsity from $0$ to $90\%$ at intervals of $10\%$ on CIFAR10/100-LT (\text{IF} = 100, $\alpha=0.5$). 
As shown in Fig.~\ref{fig:norm_sparse_visual_ablation} (b) and (c), minor sparsification yields a slight model performance enhancement, and the performance is obtained around $60\%$ sparsity. Surprisingly, a large sparsity ratio still remains superior compared to models without sparsification.

\section{Conclusion}
This paper presented \texttt{FedLoGe}, a model training framework that enhances the performance of both local and generic models in Fed-LT settings in the unified perspective of neural collapse.
The proposed framework is comprised of SSE-C, a component developed inspired by the feature collapse phenomenon to enhance representation learning, and GLA-FR, which enables fast adaptive feature realignment for both global and local models. 
As a result, \texttt{FedLoGe} attains significant performance gains over current methods in personalized and long-tail federated learning.
Future research will explore adaptive sparsity and expand the framework to diverse loss functions and tasks.

\section*{Acknowledgements}
This work is supported by the National Natural Science Foundation of China (Grant No. 62106222, No. 62201504), the Natural Science Foundation of Zhejiang Province, China(Grant No. LZ23F020008, No. LGJ22F010001), Zhejiang Lab Open Research Project (No. K2022PD0AB05) and the Zhejiang University-Angelalign Inc. R$\&$D Center for Intelligent Healthcare.

\bibliography{iclr2024_conference}

\begin{thebibliography}{72}
\providecommand{\natexlab}[1]{#1}
\providecommand{\url}[1]{\texttt{#1}}
\expandafter\ifx\csname urlstyle\endcsname\relax
  \providecommand{\doi}[1]{doi: #1}\else
  \providecommand{\doi}{doi: \begingroup \urlstyle{rm}\Url}\fi

\bibitem[Arivazhagan et~al.(2019)Arivazhagan, Aggarwal, Singh, and
  Choudhary]{arivazhagan2019federated}
Manoj~Ghuhan Arivazhagan, Vinay Aggarwal, Aaditya~Kumar Singh, and Sunav
  Choudhary.
\newblock Federated learning with personalization layers.
\newblock \emph{arXiv preprint arXiv:1912.00818}, 2019.

\bibitem[Awoyemi et~al.(2017)Awoyemi, Adetunmbi, and
  Oluwadare]{awoyemi2017credit}
John~O Awoyemi, Adebayo~O Adetunmbi, and Samuel~A Oluwadare.
\newblock Credit card fraud detection using machine learning techniques: A
  comparative analysis.
\newblock In \emph{2017 international conference on computing networking and
  informatics (ICCNI)}, pp.\  1--9. IEEE, 2017.

\bibitem[Briggs et~al.(2020)Briggs, Fan, and Andras]{briggs2020federated}
Christopher Briggs, Zhong Fan, and Peter Andras.
\newblock Federated learning with hierarchical clustering of local updates to
  improve training on non-iid data.
\newblock In \emph{2020 International Joint Conference on Neural Networks
  (IJCNN)}, pp.\  1--9. IEEE, 2020.

\bibitem[Cao et~al.(2019)Cao, Wei, Gaidon, Arechiga, and Ma]{cao2019learning}
Kaidi Cao, Colin Wei, Adrien Gaidon, Nikos Arechiga, and Tengyu Ma.
\newblock Learning imbalanced datasets with label-distribution-aware margin
  loss.
\newblock \emph{Advances in neural information processing systems}, 32, 2019.

\bibitem[Chen \& Chao(2022)Chen and Chao]{chen2022on}
Hong-You Chen and Wei-Lun Chao.
\newblock On bridging generic and personalized federated learning for image
  classification.
\newblock In \emph{International Conference on Learning Representations}, 2022.
\newblock URL \url{https://openreview.net/forum?id=I1hQbx10Kxn}.

\bibitem[Chen et~al.(2022{\natexlab{a}})Chen, Yang, Zhu, Peng, and
  Yuan]{chen2022personalized}
Zhen Chen, Chen Yang, Meilu Zhu, Zhe Peng, and Yixuan Yuan.
\newblock Personalized retrogress-resilient federated learning toward
  imbalanced medical data.
\newblock \emph{IEEE Transactions on Medical Imaging}, 41\penalty0
  (12):\penalty0 3663--3674, 2022{\natexlab{a}}.

\bibitem[Chen et~al.(2022{\natexlab{b}})Chen, Liu, Wang, Yang, Quek, and
  Liu]{chen2022towards}
Zihan Chen, Songshang Liu, Hualiang Wang, Howard~H Yang, Tony~QS Quek, and
  Zuozhu Liu.
\newblock Towards federated long-tailed learning.
\newblock \emph{arXiv preprint arXiv:2206.14988}, 2022{\natexlab{b}}.

\bibitem[Chen et~al.(2023)Chen, Yang, Quek, and Chong]{chen2023spectral}
Zihan Chen, Howard~Hao Yang, Tony Quek, and Kai Fong~Ernest Chong.
\newblock Spectral co-distillation for personalized federated learning.
\newblock In \emph{Thirty-seventh Conference on Neural Information Processing
  Systems}, 2023.

\bibitem[Chou et~al.(2022)Chou, Hong, Sun, Cai, Song, and Li]{chou2022grp}
Yen-Hsiu Chou, Shenda Hong, Chenxi Sun, Derun Cai, Moxian Song, and Hongyan Li.
\newblock Grp-fed: Addressing client imbalance in federated learning via
  global-regularized personalization.
\newblock In \emph{Proceedings of the 2022 SIAM International Conference on
  Data Mining (SDM)}, pp.\  451--458. SIAM, 2022.

\bibitem[Collins et~al.(2021)Collins, Hassani, Mokhtari, and
  Shakkottai]{collins2021exploiting}
Liam Collins, Hamed Hassani, Aryan Mokhtari, and Sanjay Shakkottai.
\newblock Exploiting shared representations for personalized federated
  learning.
\newblock In \emph{International conference on machine learning}, pp.\
  2089--2099. PMLR, 2021.

\bibitem[Dai et~al.(2022)Dai, Shen, He, Tian, and Tao]{dai2022dispfl}
Rong Dai, Li~Shen, Fengxiang He, Xinmei Tian, and Dacheng Tao.
\newblock Dispfl: Towards communication-efficient personalized federated
  learning via decentralized sparse training.
\newblock \emph{arXiv preprint arXiv:2206.00187}, 2022.

\bibitem[Dai et~al.(2023)Dai, Chen, Li, Heinecke, Sun, and Xu]{dai2023tackling}
Yutong Dai, Zeyuan Chen, Junnan Li, Shelby Heinecke, Lichao Sun, and Ran Xu.
\newblock Tackling data heterogeneity in federated learning with class
  prototypes.
\newblock In \emph{Proceedings of the AAAI Conference on Artificial
  Intelligence}, volume~37, pp.\  7314--7322, 2023.

\bibitem[Dang et~al.(2023)Dang, Nguyen, Tran, Tran, and Ho]{dang2023neural}
Hien Dang, Tan Nguyen, Tho Tran, Hung Tran, and Nhat Ho.
\newblock Neural collapse in deep linear network: From balanced to imbalanced
  data.
\newblock \emph{arXiv preprint arXiv:2301.00437}, 2023.

\bibitem[Dastile et~al.(2020)Dastile, Celik, and
  Potsane]{dastile2020statistical}
Xolani Dastile, Turgay Celik, and Moshe Potsane.
\newblock Statistical and machine learning models in credit scoring: A
  systematic literature survey.
\newblock \emph{Applied Soft Computing}, 91:\penalty0 106263, 2020.

\bibitem[Deng et~al.(2020)Deng, Kamani, and Mahdavi]{deng2020adaptive}
Yuyang Deng, Mohammad~Mahdi Kamani, and Mehrdad Mahdavi.
\newblock Adaptive personalized federated learning.
\newblock \emph{arXiv preprint arXiv:2003.13461}, 2020.

\bibitem[Diao et~al.(2020)Diao, Ding, and Tarokh]{diao2020heterofl}
Enmao Diao, Jie Ding, and Vahid Tarokh.
\newblock Heterofl: Computation and communication efficient federated learning
  for heterogeneous clients.
\newblock \emph{arXiv preprint arXiv:2010.01264}, 2020.

\bibitem[Dong et~al.(2022)Dong, Zhang, Li, and Kung]{dong2022spherefed}
Xin Dong, Sai~Qian Zhang, Ang Li, and HT~Kung.
\newblock Spherefed: Hyperspherical federated learning.
\newblock In \emph{European Conference on Computer Vision}, pp.\  165--184.
  Springer, 2022.

\bibitem[Elbatel et~al.(2023)Elbatel, Wang, Mart{\'\i}, Fu, and
  Li]{elbatel2023federated}
Marawan Elbatel, Hualiang Wang, Robert Mart{\'\i}, Huazhu Fu, and Xiaomeng Li.
\newblock Federated model aggregation via self-supervised priors for highly
  imbalanced medical image classification.
\newblock \emph{arXiv preprint arXiv:2307.14959}, 2023.

\bibitem[Fallah et~al.(2020)Fallah, Mokhtari, and
  Ozdaglar]{fallah2020personalized}
Alireza Fallah, Aryan Mokhtari, and Asuman Ozdaglar.
\newblock Personalized federated learning with theoretical guarantees: A
  model-agnostic meta-learning approach.
\newblock \emph{Advances in Neural Information Processing Systems},
  33:\penalty0 3557--3568, 2020.

\bibitem[Fang et~al.(2021)Fang, He, Long, and Su]{fang2021exploring}
Cong Fang, Hangfeng He, Qi~Long, and Weijie~J Su.
\newblock Exploring deep neural networks via layer-peeled model: Minority
  collapse in imbalanced training.
\newblock \emph{Proceedings of the National Academy of Sciences}, 118\penalty0
  (43):\penalty0 e2103091118, 2021.

\bibitem[Frankle \& Carbin(2019)Frankle and Carbin]{frankle2018the}
Jonathan Frankle and Michael Carbin.
\newblock The lottery ticket hypothesis: Finding sparse, trainable neural
  networks.
\newblock In \emph{International Conference on Learning Representations}, 2019.
\newblock URL \url{https://openreview.net/forum?id=rJl-b3RcF7}.

\bibitem[Ghosh et~al.(2020)Ghosh, Chung, Yin, and
  Ramchandran]{ghosh2020efficient}
Avishek Ghosh, Jichan Chung, Dong Yin, and Kannan Ramchandran.
\newblock An efficient framework for clustered federated learning.
\newblock \emph{Advances in Neural Information Processing Systems},
  33:\penalty0 19586--19597, 2020.

\bibitem[Glorot et~al.(2011)Glorot, Bordes, and Bengio]{glorot2011deep}
Xavier Glorot, Antoine Bordes, and Yoshua Bengio.
\newblock Deep sparse rectifier neural networks.
\newblock In \emph{Proceedings of the fourteenth international conference on
  artificial intelligence and statistics}, pp.\  315--323. JMLR Workshop and
  Conference Proceedings, 2011.

\bibitem[Huang et~al.(2021)Huang, Gupta, Song, Li, and
  Arora]{huang2021evaluating}
Yangsibo Huang, Samyak Gupta, Zhao Song, Kai Li, and Sanjeev Arora.
\newblock Evaluating gradient inversion attacks and defenses in federated
  learning.
\newblock \emph{Advances in Neural Information Processing Systems},
  34:\penalty0 7232--7241, 2021.

\bibitem[Ji et~al.(2021)Ji, Lu, Zhang, Deng, and Su]{ji2021unconstrained}
Wenlong Ji, Yiping Lu, Yiliang Zhang, Zhun Deng, and Weijie~J Su.
\newblock An unconstrained layer-peeled perspective on neural collapse.
\newblock \emph{arXiv preprint arXiv:2110.02796}, 2021.

\bibitem[Kairouz et~al.(2021)Kairouz, McMahan, Avent, Bellet, Bennis, Bhagoji,
  Bonawitz, Charles, Cormode, Cummings, et~al.]{kairouz2021advances}
Peter Kairouz, H~Brendan McMahan, Brendan Avent, Aur{\'e}lien Bellet, Mehdi
  Bennis, Arjun~Nitin Bhagoji, Kallista Bonawitz, Zachary Charles, Graham
  Cormode, Rachel Cummings, et~al.
\newblock Advances and open problems in federated learning.
\newblock \emph{Foundations and Trends{\textregistered} in Machine Learning},
  14\penalty0 (1--2):\penalty0 1--210, 2021.

\bibitem[Kang et~al.(2019)Kang, Xie, Rohrbach, Yan, Gordo, Feng, and
  Kalantidis]{kang2019decoupling}
Bingyi Kang, Saining Xie, Marcus Rohrbach, Zhicheng Yan, Albert Gordo, Jiashi
  Feng, and Yannis Kalantidis.
\newblock Decoupling representation and classifier for long-tailed recognition.
\newblock \emph{arXiv preprint arXiv:1910.09217}, 2019.

\bibitem[Kim \& Kim(2020)Kim and Kim]{kim2020adjusting}
Byungju Kim and Junmo Kim.
\newblock Adjusting decision boundary for class imbalanced learning.
\newblock \emph{IEEE Access}, 8:\penalty0 81674--81685, 2020.

\bibitem[Kim et~al.(2022)Kim, Wang, Sclaroff, and Saenko]{kim2022broad}
Donghyun Kim, Kaihong Wang, Stan Sclaroff, and Kate Saenko.
\newblock A broad study of pre-training for domain generalization and
  adaptation.
\newblock In \emph{European Conference on Computer Vision}, pp.\  621--638.
  Springer, 2022.

\bibitem[Lee \& Shin(2020)Lee and Shin]{lee2020federated}
Geun~Hyeong Lee and Soo-Yong Shin.
\newblock Federated learning on clinical benchmark data: performance
  assessment.
\newblock \emph{Journal of medical Internet research}, 22\penalty0
  (10):\penalty0 e20891, 2020.

\bibitem[Li \& Wang(2019)Li and Wang]{li2019fedmd}
Daliang Li and Junpu Wang.
\newblock Fedmd: Heterogenous federated learning via model distillation.
\newblock \emph{arXiv preprint arXiv:1910.03581}, 2019.

\bibitem[Li et~al.(2022{\natexlab{a}})Li, Cheung, and Lu]{li2022long}
Mengke Li, Yiu-ming Cheung, and Yang Lu.
\newblock Long-tailed visual recognition via gaussian clouded logit adjustment.
\newblock In \emph{Proceedings of the IEEE/CVF Conference on Computer Vision
  and Pattern Recognition}, pp.\  6929--6938, 2022{\natexlab{a}}.

\bibitem[Li et~al.(2020{\natexlab{a}})Li, Sahu, Talwalkar, and
  Smith]{li2020federated}
Tian Li, Anit~Kumar Sahu, Ameet Talwalkar, and Virginia Smith.
\newblock Federated learning: Challenges, methods, and future directions.
\newblock \emph{IEEE signal processing magazine}, 37\penalty0 (3):\penalty0
  50--60, 2020{\natexlab{a}}.

\bibitem[Li et~al.(2021{\natexlab{a}})Li, Hu, Beirami, and Smith]{li2021ditto}
Tian Li, Shengyuan Hu, Ahmad Beirami, and Virginia Smith.
\newblock Ditto: Fair and robust federated learning through personalization.
\newblock In \emph{International Conference on Machine Learning}, pp.\
  6357--6368. PMLR, 2021{\natexlab{a}}.

\bibitem[Li et~al.(2022{\natexlab{b}})Li, Liu, Zhou, Lu, Fernandez-Granda, Zhu,
  and Qu]{li2022principled}
Xiao Li, Sheng Liu, Jinxin Zhou, Xinyu Lu, Carlos Fernandez-Granda, Zhihui Zhu,
  and Qing Qu.
\newblock Principled and efficient transfer learning of deep models via neural
  collapse.
\newblock \emph{arXiv preprint arXiv:2212.12206}, 2022{\natexlab{b}}.

\bibitem[Li et~al.(2021{\natexlab{b}})Li, Jiang, Zhang, Kamp, and
  Dou]{li2021fedbn}
Xiaoxiao Li, Meirui Jiang, Xiaofei Zhang, Michael Kamp, and Qi~Dou.
\newblock Fed{\{}bn{\}}: Federated learning on non-{\{}iid{\}} features via
  local batch normalization.
\newblock In \emph{International Conference on Learning Representations},
  2021{\natexlab{b}}.
\newblock URL \url{https://openreview.net/pdf?id=6YEQUn0QICG}.

\bibitem[Li et~al.(2020{\natexlab{b}})Li, Wang, Kang, Tang, Wang, Li, and
  Feng]{li2020overcoming}
Yu~Li, Tao Wang, Bingyi Kang, Sheng Tang, Chunfeng Wang, Jintao Li, and Jiashi
  Feng.
\newblock Overcoming classifier imbalance for long-tail object detection with
  balanced group softmax.
\newblock In \emph{Proceedings of the IEEE/CVF conference on computer vision
  and pattern recognition}, pp.\  10991--11000, 2020{\natexlab{b}}.

\bibitem[Li et~al.(2023)Li, Shang, He, Lin, and Wu]{li2023no}
Zexi Li, Xinyi Shang, Rui He, Tao Lin, and Chao Wu.
\newblock No fear of classifier biases: Neural collapse inspired federated
  learning with synthetic and fixed classifier.
\newblock \emph{arXiv preprint arXiv:2303.10058}, 2023.

\bibitem[Liu et~al.(2019)Liu, Miao, Zhan, Wang, Gong, and Yu]{liu2019large}
Ziwei Liu, Zhongqi Miao, Xiaohang Zhan, Jiayun Wang, Boqing Gong, and Stella~X
  Yu.
\newblock Large-scale long-tailed recognition in an open world.
\newblock In \emph{Proceedings of the IEEE/CVF Conference on Computer Vision
  and Pattern Recognition}, pp.\  2537--2546, 2019.

\bibitem[Ma et~al.(2023)Ma, Niu, Xu, Huang, Han, and Chang]{ma2023digeo}
Jiawei Ma, Yulei Niu, Jincheng Xu, Shiyuan Huang, Guangxing Han, and Shih-Fu
  Chang.
\newblock Digeo: Discriminative geometry-aware learning for generalized
  few-shot object detection.
\newblock In \emph{Proceedings of the IEEE/CVF Conference on Computer Vision
  and Pattern Recognition}, pp.\  3208--3218, 2023.

\bibitem[McMahan et~al.(2017)McMahan, Moore, Ramage, Hampson, and
  y~Arcas]{mcmahan2017communication}
Brendan McMahan, Eider Moore, Daniel Ramage, Seth Hampson, and Blaise~Aguera
  y~Arcas.
\newblock Communication-efficient learning of deep networks from decentralized
  data.
\newblock In \emph{Artificial intelligence and statistics}, pp.\  1273--1282.
  PMLR, 2017.

\bibitem[Mixon et~al.(2020)Mixon, Parshall, and Pi]{mixon2020neural}
Dustin~G Mixon, Hans Parshall, and Jianzong Pi.
\newblock Neural collapse with unconstrained features.
\newblock \emph{arXiv preprint arXiv:2011.11619}, 2020.

\bibitem[Oh et~al.(2022)Oh, Kim, and Yun]{oh2022fedbabu}
Jaehoon Oh, SangMook Kim, and Se-Young Yun.
\newblock Fed{BABU}: Toward enhanced representation for federated image
  classification.
\newblock In \emph{International Conference on Learning Representations}, 2022.
\newblock URL \url{https://openreview.net/forum?id=HuaYQfggn5u}.

\bibitem[Papyan et~al.(2020)Papyan, Han, and Donoho]{papyan2020prevalence}
Vardan Papyan, XY~Han, and David~L Donoho.
\newblock Prevalence of neural collapse during the terminal phase of deep
  learning training.
\newblock \emph{Proceedings of the National Academy of Sciences}, 117\penalty0
  (40):\penalty0 24652--24663, 2020.

\bibitem[Pillutla et~al.(2022)Pillutla, Malik, Mohamed, Rabbat, Sanjabi, and
  Xiao]{pillutla2022federated}
Krishna Pillutla, Kshitiz Malik, Abdel-Rahman Mohamed, Mike Rabbat, Maziar
  Sanjabi, and Lin Xiao.
\newblock Federated learning with partial model personalization.
\newblock In \emph{International Conference on Machine Learning}, pp.\
  17716--17758. PMLR, 2022.

\bibitem[Qian et~al.(2023)Qian, Lu, and Wang]{qian2023long}
Pinxin Qian, Yang Lu, and Hanzi Wang.
\newblock Long-tailed federated learning via aggregated meta mapping.
\newblock In \emph{2023 IEEE International Conference on Image Processing
  (ICIP)}, pp.\  2010--2014. IEEE, 2023.

\bibitem[Sattler et~al.(2020)Sattler, M{\"u}ller, and
  Samek]{sattler2020clustered}
Felix Sattler, Klaus-Robert M{\"u}ller, and Wojciech Samek.
\newblock Clustered federated learning: Model-agnostic distributed multitask
  optimization under privacy constraints.
\newblock \emph{IEEE transactions on neural networks and learning systems},
  32\penalty0 (8):\penalty0 3710--3722, 2020.

\bibitem[Shang et~al.(2022{\natexlab{a}})Shang, Lu, Cheung, and
  Wang]{shang2022fedic}
Xinyi Shang, Yang Lu, Yiu-ming Cheung, and Hanzi Wang.
\newblock Fedic: Federated learning on non-iid and long-tailed data via
  calibrated distillation.
\newblock In \emph{2022 IEEE International Conference on Multimedia and Expo
  (ICME)}, pp.\  1--6. IEEE, 2022{\natexlab{a}}.

\bibitem[Shang et~al.(2022{\natexlab{b}})Shang, Lu, Huang, and
  Wang]{shang2022federated}
Xinyi Shang, Yang Lu, Gang Huang, and Hanzi Wang.
\newblock Federated learning on heterogeneous and long-tailed data via
  classifier re-training with federated features.
\newblock \emph{arXiv preprint arXiv:2204.13399}, 2022{\natexlab{b}}.

\bibitem[Shen et~al.(2022)Shen, Zhou, and Yu]{shen2022cd2}
Yiqing Shen, Yuyin Zhou, and Lequan Yu.
\newblock Cd2-pfed: Cyclic distillation-guided channel decoupling for model
  personalization in federated learning.
\newblock In \emph{Proceedings of the IEEE/CVF Conference on Computer Vision
  and Pattern Recognition}, pp.\  10041--10050, 2022.

\bibitem[Shen et~al.(2021)Shen, Cervino, Hassani, and
  Ribeiro]{shen2021agnostic}
Zebang Shen, Juan Cervino, Hamed Hassani, and Alejandro Ribeiro.
\newblock An agnostic approach to federated learning with class imbalance.
\newblock In \emph{International Conference on Learning Representations}, 2021.

\bibitem[T~Dinh et~al.(2020)T~Dinh, Tran, and Nguyen]{t2020personalized}
Canh T~Dinh, Nguyen Tran, and Josh Nguyen.
\newblock Personalized federated learning with moreau envelopes.
\newblock \emph{Advances in Neural Information Processing Systems},
  33:\penalty0 21394--21405, 2020.

\bibitem[Tan et~al.(2022)Tan, Yu, Cui, and Yang]{tan2022towards}
Alysa~Ziying Tan, Han Yu, Lizhen Cui, and Qiang Yang.
\newblock Towards personalized federated learning.
\newblock \emph{IEEE Transactions on Neural Networks and Learning Systems},
  2022.

\bibitem[Tan et~al.(2021)Tan, Lu, Zhang, Yin, and Li]{tan2021equalization}
Jingru Tan, Xin Lu, Gang Zhang, Changqing Yin, and Quanquan Li.
\newblock Equalization loss v2: A new gradient balance approach for long-tailed
  object detection.
\newblock In \emph{Proceedings of the IEEE/CVF conference on computer vision
  and pattern recognition}, pp.\  1685--1694, 2021.

\bibitem[Thrampoulidis et~al.(2022)Thrampoulidis, Kini, Vakilian, and
  Behnia]{thrampoulidis2022imbalance}
Christos Thrampoulidis, Ganesh~Ramachandra Kini, Vala Vakilian, and Tina
  Behnia.
\newblock Imbalance trouble: Revisiting neural-collapse geometry.
\newblock \emph{Advances in Neural Information Processing Systems},
  35:\penalty0 27225--27238, 2022.

\bibitem[Tirer \& Bruna(2022)Tirer and Bruna]{tirer2022extended}
Tom Tirer and Joan Bruna.
\newblock Extended unconstrained features model for exploring deep neural
  collapse.
\newblock In \emph{International Conference on Machine Learning}, pp.\
  21478--21505. PMLR, 2022.

\bibitem[Van~Horn et~al.(2018)Van~Horn, Mac~Aodha, Song, Cui, Sun, Shepard,
  Adam, Perona, and Belongie]{van2018inaturalist}
Grant Van~Horn, Oisin Mac~Aodha, Yang Song, Yin Cui, Chen Sun, Alex Shepard,
  Hartwig Adam, Pietro Perona, and Serge Belongie.
\newblock The inaturalist species classification and detection dataset.
\newblock In \emph{Proceedings of the IEEE conference on computer vision and
  pattern recognition}, pp.\  8769--8778, 2018.

\bibitem[Vasconcelos et~al.(2022)Vasconcelos, Birodkar, and
  Dumoulin]{vasconcelos2022proper}
Cristina Vasconcelos, Vighnesh Birodkar, and Vincent Dumoulin.
\newblock Proper reuse of image classification features improves object
  detection.
\newblock In \emph{Proceedings of the IEEE/CVF conference on computer vision
  and pattern recognition}, pp.\  13628--13637, 2022.

\bibitem[Wang et~al.(2022)Wang, Wang, and Shen]{wang2022logit}
Huan Wang, Lijuan Wang, and Jun Shen.
\newblock Logit calibration for non-iid and long-tailed data in federated
  learning.
\newblock In \emph{2022 IEEE Intl Conf on Parallel \& Distributed Processing
  with Applications, Big Data \& Cloud Computing, Sustainable Computing \&
  Communications, Social Computing \& Networking
  (ISPA/BDCloud/SocialCom/SustainCom)}, pp.\  782--789. IEEE, 2022.

\bibitem[Wang et~al.(2021)Wang, Xu, Wang, and Zhu]{wang2021addressing}
Lixu Wang, Shichao Xu, Xiao Wang, and Qi~Zhu.
\newblock Addressing class imbalance in federated learning.
\newblock In \emph{Proceedings of the AAAI Conference on Artificial
  Intelligence}, volume~35, pp.\  10165--10173, 2021.

\bibitem[Wang et~al.(2020)Wang, Xiang, Zou, and Xu]{wang2020mma}
Zhennan Wang, Canqun Xiang, Wenbin Zou, and Chen Xu.
\newblock Mma regularization: Decorrelating weights of neural networks by
  maximizing the minimal angles.
\newblock \emph{Advances in Neural Information Processing Systems},
  33:\penalty0 19099--19110, 2020.

\bibitem[Xiao et~al.(2023)Xiao, Chen, Liu, Wang, FENG, Hao, Zhou, Wu, Yang, and
  Liu]{xiao2023fedgrab}
Zikai Xiao, Zihan Chen, Songshang Liu, Hualiang Wang, YANG FENG, Jin Hao,
  Joey~Tianyi Zhou, Jian Wu, Howard~Hao Yang, and Zuozhu Liu.
\newblock Fed-grab: Federated long-tailed learning with self-adjusting gradient
  balancer.
\newblock In \emph{Thirty-seventh Conference on Neural Information Processing
  Systems}, 2023.

\bibitem[Yang et~al.(2021)Yang, Wang, Zhu, Wang, and Qian]{yang2021federated}
Miao Yang, Ximin Wang, Hongbin Zhu, Haifeng Wang, and Hua Qian.
\newblock Federated learning with class imbalance reduction.
\newblock In \emph{2021 29th European Signal Processing Conference (EUSIPCO)},
  pp.\  2174--2178. IEEE, 2021.

\bibitem[Yang et~al.(2023{\natexlab{a}})Yang, Chen, Zhao, Meng, Zhou, and
  Sun]{yang2023integrating}
Wenkai Yang, Deli Chen, Hao Zhao, Fandong Meng, Jie Zhou, and Xu~Sun.
\newblock Integrating local real data with global gradient prototypes for
  classifier re-balancing in federated long-tailed learning.
\newblock \emph{arXiv preprint arXiv:2301.10394}, 2023{\natexlab{a}}.

\bibitem[Yang et~al.(2022)Yang, Chen, Li, Xie, Lin, and Tao]{yang2022inducing}
Yibo Yang, Shixiang Chen, Xiangtai Li, Liang Xie, Zhouchen Lin, and Dacheng
  Tao.
\newblock Inducing neural collapse in imbalanced learning: Do we really need a
  learnable classifier at the end of deep neural network?
\newblock \emph{Advances in Neural Information Processing Systems},
  35:\penalty0 37991--38002, 2022.

\bibitem[Yang et~al.(2023{\natexlab{b}})Yang, Yuan, Li, Lin, Torr, and
  Tao]{yang2023neural}
Yibo Yang, Haobo Yuan, Xiangtai Li, Zhouchen Lin, Philip Torr, and Dacheng Tao.
\newblock Neural collapse inspired feature-classifier alignment for few-shot
  class incremental learning.
\newblock \emph{arXiv preprint arXiv:2302.03004}, 2023{\natexlab{b}}.

\bibitem[Yaras et~al.(2022)Yaras, Wang, Zhu, Balzano, and Qu]{yaras2022neural}
Can Yaras, Peng Wang, Zhihui Zhu, Laura Balzano, and Qing Qu.
\newblock Neural collapse with normalized features: A geometric analysis over
  the riemannian manifold.
\newblock \emph{Advances in neural information processing systems},
  35:\penalty0 11547--11560, 2022.

\bibitem[Zhang et~al.(2023)Zhang, Li, Tang, Sun, Chen, Zhang, Chen, Chen, and
  Li]{zhang2023fed}
Jianyi Zhang, Ang Li, Minxue Tang, Jingwei Sun, Xiang Chen, Fan Zhang, Changyou
  Chen, Yiran Chen, and Hai Li.
\newblock Fed-cbs: A heterogeneity-aware client sampling mechanism for
  federated learning via class-imbalance reduction.
\newblock In \emph{International Conference on Machine Learning}, pp.\
  41354--41381. PMLR, 2023.

\bibitem[Zhang et~al.(2022{\natexlab{a}})Zhang, Li, Li, Xu, Wu, Ding, and
  Wu]{zhang2022federated}
Jie Zhang, Zhiqi Li, Bo~Li, Jianghe Xu, Shuang Wu, Shouhong Ding, and Chao Wu.
\newblock Federated learning with label distribution skew via logits
  calibration.
\newblock In \emph{International Conference on Machine Learning}, pp.\
  26311--26329. PMLR, 2022{\natexlab{a}}.

\bibitem[Zhang et~al.(2022{\natexlab{b}})Zhang, Li, Li, Guo, and
  Shao]{zhang2022personalized}
Xu~Zhang, Yinchuan Li, Wenpeng Li, Kaiyang Guo, and Yunfeng Shao.
\newblock Personalized federated learning via variational bayesian inference.
\newblock In \emph{International Conference on Machine Learning}, pp.\
  26293--26310. PMLR, 2022{\natexlab{b}}.

\bibitem[Zhong et~al.(2023)Zhong, Cui, Yang, Wu, Qi, Zhang, and
  Jia]{zhong2023understanding}
Zhisheng Zhong, Jiequan Cui, Yibo Yang, Xiaoyang Wu, Xiaojuan Qi, Xiangyu
  Zhang, and Jiaya Jia.
\newblock Understanding imbalanced semantic segmentation through neural
  collapse.
\newblock In \emph{Proceedings of the IEEE/CVF Conference on Computer Vision
  and Pattern Recognition}, pp.\  19550--19560, 2023.

\bibitem[Zhu et~al.(2021)Zhu, Ding, Zhou, Li, You, Sulam, and
  Qu]{zhu2021geometric}
Zhihui Zhu, Tianyu Ding, Jinxin Zhou, Xiao Li, Chong You, Jeremias Sulam, and
  Qing Qu.
\newblock A geometric analysis of neural collapse with unconstrained features.
\newblock \emph{Advances in Neural Information Processing Systems},
  34:\penalty0 29820--29834, 2021.

\end{thebibliography}
\bibliographystyle{iclr2024_conference}

\appendix
\section{Appendix}

\subsection{Detailed setup}
\noindent\textbf{Dataset, Models and Metrics:}
We consider image classification tasks for performance evaluation on benchmark long-tailed datasets: CIFAR-10/100-LT, ImageNet-LT, and iNaturalist-User-160k. The CIFAR-10/100-LT datasets are sampled into a long-tailed distribution employing an exponential distribution governed by Imbalance Factor ($\text{IF}$) in \cite{cao2019learning}, calculated over the comprehensive dataset $\mathcal{D}$. Consistency is maintained by utilizing configurations from \cite{liu2019large} for ImageNet-LT, where the number of images per class varies between 5 and 1280. Additionally, to evaluate performance with real-world data, the study includes experiments on iNaturalist-User-160k, comprising 160k examples across 1023 species classes, sampled from iNaturalist-2017 in \cite{van2018inaturalist}.

All experiments employ non-IID data partitions, achieved through Dirichlet distributions. The concentration parameter $\alpha$ serves to regulate the identicalness of local data distributions among all clients. A ResNet-18 is trained over $K=40$ clients on CIFAR-10-LT, while ResNet-34 and ResNet-50 are implemented on CIFAR-100-LT and ImageNet-LT, respectively, with $N=20$ clients. The configurations for iNaturalist-160k align with those utilized for ImageNet-LT.

The balanced dataset is utilized for the evaluation of global model performance. The global accuracy is calculated as the ratio of the number of correct predictions to the total number of predictions. We also report the accuracy across many, medium, and few classes. The categorization process unfolds as follows: Initially, all classes are sorted based on the number of data samples in descending order. Subsequently, two thresholds (e.g., $75\%$, $95\%$) are set to segregate the classes into head classes, middle classes, and tail classes.

For CIFAR10/100-LT datasets, we engaged 40 clients with full participation in each training round, setting the number of local epochs to 5. For ImageNet/iNaturalist, the experiments involved 20 clients, with a $40\%$ participation rate in each training round, and the number of local epochs was set to 3.

When assessing personalized models, the test set is sampled from the global balanced dataset by local distribution. We ensure that the distribution of the local train set and the local test set is perfectly aligned. The accuracy of local models is the arithmetic mean of all clients.

\begin{figure}[htbp]
\centering
\includegraphics[width=0.5\textwidth]{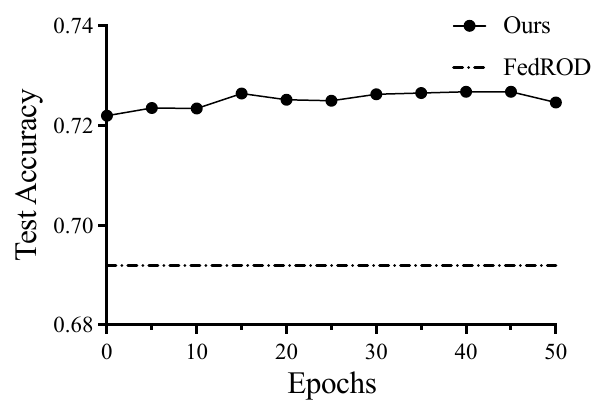}
\caption{Accuracy of local training after GLA-FR.}
\label{fig:local_finetune}
\end{figure}

\noindent\textbf{Training Settings:}
We train the generic model with $T=500$ rounds via applying SGD optimizer for local training in all experiments unless otherwise stated. 
We use $\alpha=1, 0.5$ and $\text{IF}$ = $50, 100$ in CIFAR-10/100-LT.  We use $\alpha= 0.1, 0.5$ in ImageNet-LT and iNaturalist, respectively. 


\subsection{Table of Notations}

Please refer to Table~\ref{tab:notations} for the table of notations used throughout this paper.

\begin{table}[h!]
\centering
\begin{tabular}{@{}cl@{}}
\toprule
\textbf{Notation} & \multicolumn{1}{c}{\textbf{Description}} \\ \midrule
$K$               & Number of clients in the federated learning system. \\
$w$               & Parameters of the generic model, consisting of $\theta$ and $\mathbf{\psi}$. \\
$w_k$             & Parameters of the $k^{\text{th}}$ personalized model for client $k$. \\
$f(x, \theta)$    & Feature extractor parameterized by $\theta$. \\
$h$               & Features transformed from input $x$ by feature extractor $f$. \\
$g(h, \mathbf{\psi})$ & Generic classifier mapping features $h$ to output labels. \\
$g(h, \mathbf{\phi}_k)$ & Personalized classifier for client $k$. \\
$\mathbf{\phi}_k$ & Personalized classifier parameters for client $k$. \\
$\mathcal{D}$     & Global long-tailed dataset. \\
$\mathcal{D}_k$   & Local dataset of client $k$. \\
$T$               & Number of training rounds. \\
$c$               & Class index. \\
$\mathbf{\psi}_c$ & Classifier vector corresponding to class $c$. \\
$h_c$             & Features corresponding to class $c$. \\
$\mathbf{S}$      & Sparse indicator matrix. \\
$\gamma$          & Predetermined $\ell_2$ norm value for class vectors. \\
$\mathcal{L}_{\text{SSE-C}}$ & Loss function for Static Sparse Equiangular Tight Frame Classifier (SSE-C). \\
$\mathcal{L}_{\text{norm}}$, $\mathcal{L}_{\text{angle}}$ & Loss functions for optimizing the geometric structure of SSE-C. \\
$n_c$             & Sample number of class $c$. \\
$\mu_c$           & Class mean of features for class $c$. \\
$h_{i,c}$         & Feature of class $c$ for sample $i$. \\ \bottomrule
\end{tabular}
\caption{Summary of notations used in the paper.}
\label{tab:notations}
\end{table}

\subsection{Detailes of feature degeneration}

Given that $n_c$ is the sample number of class $c$.  $\mu_c$ is the class mean of class $c$, and $h_{i,c}$ is the feature of class $c$ on sample $i$, the features $h_{i,c}$ will collapse to the within-class mean $\mu_c = \frac{1}{n_c} \sum_{i=1}^{n_c}h_{i,c}$. This indicates that the covariance $\frac{1}{n_c} \sum_{i=1}^{n_c}(h_{i,c}-\mu_c)(h_{i,c}-\mu_c)^T$ will converge to $0$. We investigate whether $h_{i,c} - \mu_c$ tends toward zero when training with a fixed ETF classifier, as shown in Fig. ~\ref{fig:mean_std}. The steps to obtain the data in Fig. ~\ref{fig:mean_std} are: 1. Select class $c$. 2. During global test, compute all sample features $h_{i,c}$, representing the number of features. Use $h_{i,c}[d]$ to denote the $d^{th}$ feature in $h_{i,c}$. 3. Compute the mean and relative variance of $h_{c}$, sorting them in descending order of the mean, and accordingly adjust the variance positions. 
It becomes evident that not all features, $h_{i,c}[d] - \mu_c[d]$, converge to zero, and smaller $\mu_c[d]$ values exhibit greater convergence noise. The process illustrated in Fig. ~\ref{fig:mean_std} (b) follows the same steps as Fig. ~\ref{fig:mean_std} (a), but with grey vertical shadows for the sparsely positioned features. It can be observed that sparse ETF training effectively squeezes poor-quality and noisy features into these sparsely allocated positions.

\subsection{Computational Cost}
We evaluate the computational expense of \texttt{Fed-LoGe} across its three stages. In the initial representation stage, the server is tasked with constructing the SSE-C. Setting the learning rate at 0.0001 and executing 10,000 optimization steps, we conducted the experiments three times on the PyTorch platform utilizing the NVIDIA GeForce RTX 3090; the average cost is 11 minutes and 52 seconds.

 \begin{table}[t!]
\vspace{-4pt}
\centering

\begin{tabular}{lc}
\hline
Method  & Time Cost/Round  \\ \hline
FedAvg  &     3min11s                          \\
FedProx  &    5min24s                        \\
FedRep  &     4min07s                      \\
FedBABU &     3min14s                    \\
FedLoGe &     3min19s                  \\ \hline
\end{tabular}

\caption{Computational cost of \text{Fed-LoGe}.}
\label{tab:ablations}
\vspace{-15pt}
\end{table}

\subsection{Local Finetune Further Improve Performance}
After the personalized feature realignment, we proceed with local training without any aggregation. On CIFAR100-LT with $\text{IF}=100$ and $\alpha=0.5$, the accuracy consistently increases with additional local epochs, reaching stability at 15 epochs, as illustrated in Fig.~\ref{fig:local_finetune}.


\subsection{T-SNE Visualization of Class Means}

\begin{figure}[htbp]
    \centering
    \scriptsize
    \begin{tabular}{cc}
        \includegraphics[width=0.42\columnwidth]{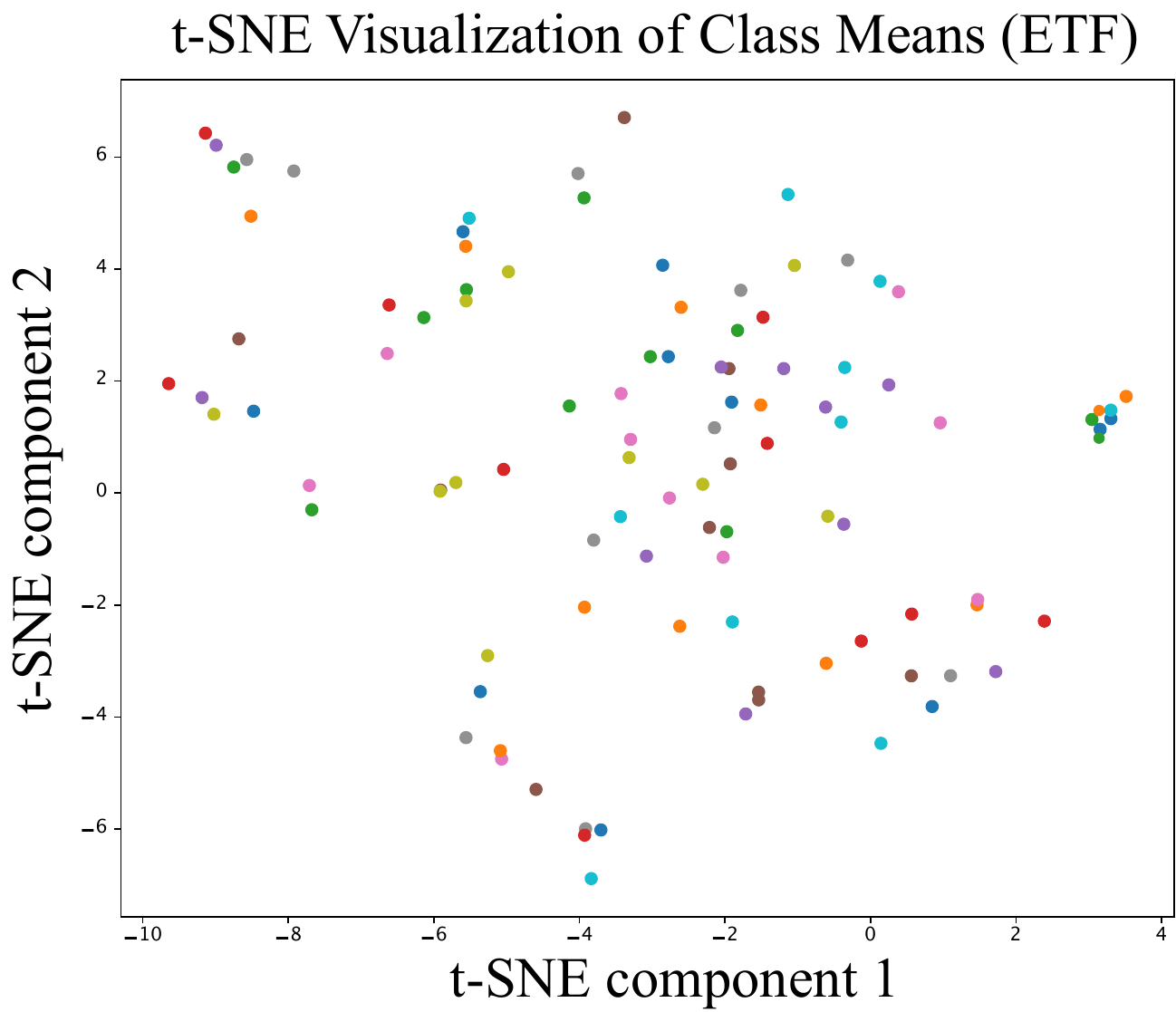}  &
        \includegraphics[width=0.42\columnwidth]{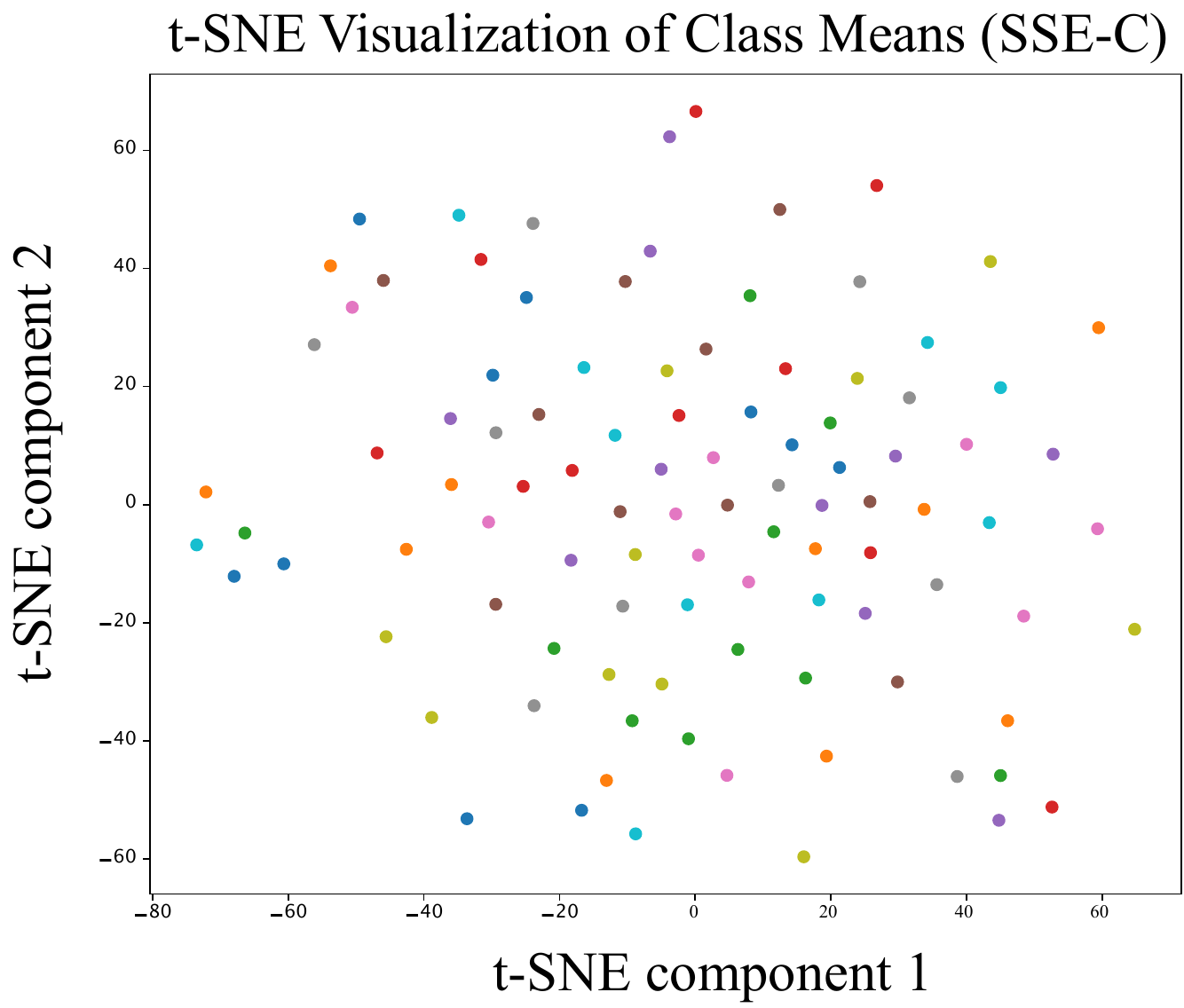}           \\
        (a) & (b)
    \end{tabular}
    \caption{t-SNE visualization of class means of ETF (dense) and SSE-C (sparse) }
    \vspace{-5pt}
    \label{fig:tsne}
\end{figure}

In this study, we compare features from models trained with SSE-C (sparse) and ETF (dense) using t-SNE, a common method for reducing dimensions, shown in Fig.~\ref{fig:tsne}. This helps us see how different training methods affect feature distribution. We notice that with SSE-C, the average features of each class are evenly spread out and quite similar in distance to each other. On the other hand, the model trained with ETF has many class averages very close or even overlapping, making the overall spread more clustered and the model more likely to make mistakes.

\subsection{Norm and Angles of SSE-C}
We have individually computed the statistical measures (mean of norm, variance of norm, mean of mutual angles, and variance of mutual angles) of the classifier vectors within SSE-C under varying sparse ratios $\beta$ and assigned norms $\gamma$, as shown in Table \ref{tab:ssec_contruct_norm_angle}. It is observable that the constructed SSE-C nearly fulfills the properties of an Equiangular Tight Frame (ETF): equal norms, and maximized, equal mutual angles. Different sparse ratios and assigned norms do not significantly influence the final construction.

\renewcommand\arraystretch{1.2}
\begin{table*}[t!]
\centering
\begin{adjustbox}{width=0.9\textwidth,center}
\begin{tabular}{cccccc}
\hline
Sparse Ratio & Norm & Mean of Norm & Var of Norm & Mean of Angle & Var of Angle \\ \hline
0.6          & 0.5  & 0.50         & 3.55e-12    & 90.06         & 0.09         \\
0.6          & 1.0  & 1.00         & 4.75e-11    & 90.06         & 0.66         \\
0.6          & 1.5  & 1.50         & 1.71e-11    & 90.04         & 1.21         \\
0.4          & 0.5  & 0.50         & 1.10e-11    & 90.06         & 0.03         \\
0.4          & 1.0  & 1.00         & 7.14e-12    & 90.06         & 0.15         \\
0.4          & 1.5  & 1.50         & 3.5e-11     & 90.06         & 0.79         \\ \hline
\end{tabular}
\end{adjustbox}
\caption{Construct measures of SSE-C }
\vspace{-10pt}
\label{tab:ssec_contruct_norm_angle}
\end{table*}

\subsection{Local Norm or local distribution for LA-FR}

Upon the backbone $\theta$, trained through $\mathbf{\psi}_{SSE-C}$, we proceed with local adaptive feature realignment. In our methodology, local norm is employed to align features with personalized preference. It is intuitive to consider that the real local distribution is more representative of the model's preference, and moreover, the local distribution is free for utilization without any computation. Consequently, we explore whether local norm or local distribution yields superior performance. 

On CIFAR100-LT with $\alpha=0.5$ and $\text{IF}=100$, we realign personalized models employing both norm and distribution, calculating the arithmetic mean, macro average, and weighted average of all clients' performance. The results can be seen in Tab. ~\ref{tab:norm_better_than_dist}. We discern that utilizing norm for realignment is more effective than employing the real distribution. One rationale is that the norm more aptly signifies the model’s cognitive capacity over the dataset. For instance, although class A encompasses more samples than class B, B is more prone to misprediction owing to its high resemblance with other classes. In this scenario, realignment ought to adhere to model cognition, fortifying class A over B.

\renewcommand\arraystretch{1.2}
\begin{table*}[t!]
\centering
\begin{adjustbox}{width=0.66\textwidth,center}
\begin{tabular}{lccc}
\hline
Model                         & PM(Avg) & PM(Macro) & PM(Weighted) \\    \hline
LA-LR (real distribution)  & 0.6200  & 0.2680    & 0.5604       \\
LA-LR (norm)               & 0.7273  & 0.5316    & 0.7129   \\   \hline
\end{tabular}
\end{adjustbox}
\caption{Local adaptive feature alignment of real distribution and personalized norm}
\vspace{-10pt}
\label{tab:norm_better_than_dist}
\end{table*}

\subsection{Projection Layer and Dot-regression Loss is Not Necessary for \text{Fed-LoGe}}
In previous work, both the projection layer and dot-regression loss are deemed essential for training the backbone (\cite{li2023no, yang2023neural}). The projection layer, being a dense structure, necessitates substantial computational cost. However, owing to the advanced design of SSE-C, Fed-LoGe does not depend on the projection layer and dot-regression loss. We undertake experiments to investigate how the dot-regression loss and projection layer influence model performance based on ETF and FedLoGe, with the results documented in Tab.\ref{tab:proj_dr_loss}. We ascertain that solely through the sparsity design of SSE-C, performance surpassing that of employing both the projection layer and dot-regression loss can be achieved.

\renewcommand\arraystretch{1.2}
\begin{table*}[t!]
\centering
\begin{adjustbox}{width=0.7\textwidth,center}
\begin{tabular}{lllll}
\hline
Method                  & Many   & Med    & Few    & All    \\ \hline
ETF                     & 0.6904 & 0.4751 & 0.1820 & 0.3825 \\
ETF+DR Loss             & 0.6882 & 0.4607 & 0.2059 & 0.3932 \\
ETF+Proj+DR Loss        & 0.6909 & 0.5011 & 0.2310 & 0.4109 \\
ETF+Proj+DR Loss+Sparse & 0.7013 & 0.5142 & 0.2479 & 0.4237 \\
FedLoGe                 & 0.7137 & 0.4989 & 0.2179 & 0.4249 \\
FedLoGe+Proj            & 0.7014 & 0.4642 & 0.1887 & 0.4004 \\ \hline
\end{tabular}
\end{adjustbox}
\caption{Ablations of projection layer and dot-regression loss.}
\vspace{-10pt}
\label{tab:proj_dr_loss}
\end{table*}

\subsection{Difference initialization methods of fixed classifier}
To further explore the effectiveness of sparsification for a fixed classifier, we have selected various classifier initialization methods (Xavier, Gaussian, Uniform, Orthogonal, Kaiming Uniform), and subsequently applied sparsification to classifiers constructed with these initializations. The results are documented in Tab.~\ref{tab:sparse_with_other_initialization}. It is observable that irrespective of the initialization method employed, post-sparsification and subsequent training of the fixed classifier yield exceptionally favorable performance.
\renewcommand\arraystretch{1.2}
\begin{table*}[t!]
\centering
\begin{adjustbox}{width=0.7\textwidth,center}
\begin{tabular}{lllll}
\hline
Initialization Method  & Many   & Med    & Few    & All    \\ \hline
Xavier                 & 0.6793 & 0.4204 & 0.1823 & 0.3784 \\
Sparse Xavier          & 0.7089 & 0.4977 & 0.2189 & 0.4237 \\
Gassian                & 0.6167 & 0.3858 & 0.1572 & 0.3407 \\
Sparse Gassian         & 0.7119 & 0.4969 & 0.2117 & 0.4209 \\
Uniform                & 0.6119 & 0.3631 & 0.1638 & 0.3366 \\
Sparse Uniform         & 0.7141 & 0.4927 & 0.2174 & 0.4231 \\
Orthogonal             & 0.6844 & 0.4446 & 0.1868 & 0.3882 \\
Sparse Orthogonal      & 0.7022 & 0.4846 & 0.2060 & 0.4124 \\
Kaiming Uniform        & 0.6933 & 0.4392 & 0.1881 & 0.3898 \\
Sparse Kaiming Uniform & 0.7130 & 0.4946 & 0.2168 & 0.4230 \\ \hline
\end{tabular}
\end{adjustbox}
\caption{Sparsification with different initialization of frozen classifiers}
\vspace{-10pt}
\label{tab:sparse_with_other_initialization}
\end{table*}

\subsection{Discussion of Privacy}

In the entire training process of FedLoGe, no extra information compared with FedAvg is transmitted: 1. During the 'Representation Learning with SSE-C' phase, FedLoGe transmits local model weights as FedAvg necessitates. 2. In the 'Feature Realignment' phase, the local models are already downloaded for norm analysis, and there are no further information-sharing operations required.

As the norm analysis is performed on the client side or global side, there would still be a few privacy concerns. However, noteworthy that the potential privacy issue exists in the general FL frameworks rather than specific to our proposed Global and Local Adaptive Feature Realignment (GLA-FR). For instance, gradient inversion~\cite{huang2021evaluating} can pose a threat to almost all gradient transmission-based FL methods without any privacy-preservation techniques. As the privacy issue of the FL framework is beyond the scope of this work.

\subsection{Convergence Experiments}
We have conducted experimental investigations into the convergence. Specifically, we explored the number of rounds required to reach an accuracy of 0.33 on CIFAR100 with an imbalance factor (IF) of 100 and $\alpha$ = 0.5, as documented in the table below. It is evident that the convergence rate of FedLoGe is almost on par with FedAvg and surpasses the current state-of-the-art (SOTA) methods.

\begin{table}[ht]
\centering
\begin{tabular}{l c}
\hline
Methods & Convergence Round \\
\hline
FedETF & 83 \\
FedROD & 68 \\
FedLoGe (Ours) & 62 \\
\hline
\end{tabular}
\caption{Comparison of Convergence Rounds for Different Methods.}
\label{tab:convergence_rounds}
\end{table}

\subsection{Comparison of Existing Methods}
\label{subsec:comparison}
We are the first to achieve joint global and local training in Federated long-tailed learning from the perspective of neural collapse. In the introduction, we analyzed the improvements of FedLoGe over some existing methods, and here we further provide the discussion:

First, most existing methods in personalized federated learning (\cite{collins2021exploiting, li2021ditto, t2020personalized, zhang2022personalized, dai2022dispfl}) utilize the global model to regularize or construct personalized models but do not evaluate them in a generic setup, failing to yield a strong global model without long-tail bias. This makes them unsuitable as a strong incentive for attracting new clients. In contrast, our approach learns models that excel in both personalized (GA-FR) and generic (FA-LR) setups without compromising either.

Second, existing methods in federated long-tailed learning (\cite{shang2022federated, yang2023integrating, shang2022fedic, wang2022logit, qian2023long}), focusing only on calibrating long-tail bias for a single global model, overlook the aspect of representation learning (backbone) under federated long-tailed settings and fail to provide personalized models. For instance, CreFF involves post-classifier retraining but does not advance in boosting the backbone and does not offer personalized models.  In contrast, Our FedLoGe achieves strong representation learning with SSE-C to support both global model and local models.

Second, existing work involving fixed classifiers has not delved deeply into the analysis of the initialization of these fixed classifiers. FedBABU (\cite{oh2022fedbabu}), for instance, merely employs a randomly initialized fixed linear layer. For FedETF (\cite{li2023no}), the Vallina fixed ETF classifier has suffered from feature degeneration, characterized by significant fluctuations between features and the class mean, while FedLoGe alleviates this issue by filtering out noisy smaller mean features (Fig.~\ref{fig:mean_std}), making features more concise and effective (Fig.~\ref{fig:pruning_experiments} (a,b) and Fig.~\ref{fig:tsne}). Besides, FedLoGe is less computationally expensive and easier to adapt to existing algorithms. DR loss is necessary for FedETF, which means some modifications are required when integrating with existing algorithms. Besides, FedETF needs a projection layer for optimization. For ResNet18 with four blocks, a 512x512 projection layer would be approximately 1MB, almost doubling the parameter size of Block 1, which is 0.56MB.

\subsection{Impact of Client Numbers and Participation Rates}

To investigate the adaptability and efficiency of FedLoGe in varied federated learning environments, we designed experiments focusing on two main variables: the number of clients and their participation rates. The experiments spanned across CIFAR10/100-LT and ImageNet/iNat datasets, tailored to simulate environments ranging from full participation to more realistic, sporadic engagement.

For CIFAR10/100-LT, we utilized a setup involving 40 clients with 100\% participation and 5 local epochs per training round. Conversely, ImageNet/iNat experiments were conducted with 20 clients at a 40\% participation rate and 3 local epochs, introducing a scenario with reduced participation and increased data complexity.

Additionally, to explore the behavior of FedLoGe under more stringent conditions, we conducted experiments on CIFAR100-LT with an imbalance factor of 100 and a heterogeneity coefficient of 0.5, decreasing the participation rate to 30\%. This setup aimed to test FedLoGe's resilience and personalized performance against data imbalance and client heterogeneity. Extending our exploration to a cross-device FL setting, the client count was increased to 100 with a 30\% participation rate, challenging FedLoGe with a highly distributed environment and testing its scalability and performance.

The experimental outcomes, detailed in Table \ref{table:performance_impact}, highlight FedLoGe's superior performance across different settings, particularly showcasing its strengths in scenarios with varied client participation rates and numbers.

\begin{table}[h!]
\centering
\begin{tabular}{lccc}
\hline
Method & 40 clients (100\% participate) & 40 clients (30\%) & 100 clients (30\%) \\
\hline
FedAvg & GM0.3818/PM0.6214 & 0.3746/0.6205 & 0.3718/0.6384 \\
FedRoD & 0.3919/0.6919 & 0.3758/0.7156 & 0.3727/0.7665 \\
FedETF & 0.3825/0.6421 & 0.4180/0.7083 & 0.3847/0.7822 \\
FedLoGe & \textbf{0.4233/0.7285} & \textbf{0.4206/0.7524} & \textbf{0.4159/0.8008} \\
\hline
\end{tabular}
\caption{Impact of client diversity and participation rates on FedLoGe performance compared to other FL algorithms.}
\label{table:performance_impact}
\end{table}

These results confirm FedLoGe's robustness and adaptability, underlining its potential for broad applicability in federated learning scenarios characterized by varying degrees of client engagement and diversity.

\subsection{Effectiveness of SSE-C}

\begin{figure}[htbp]
    \centering
    \includegraphics[width=0.3\columnwidth]{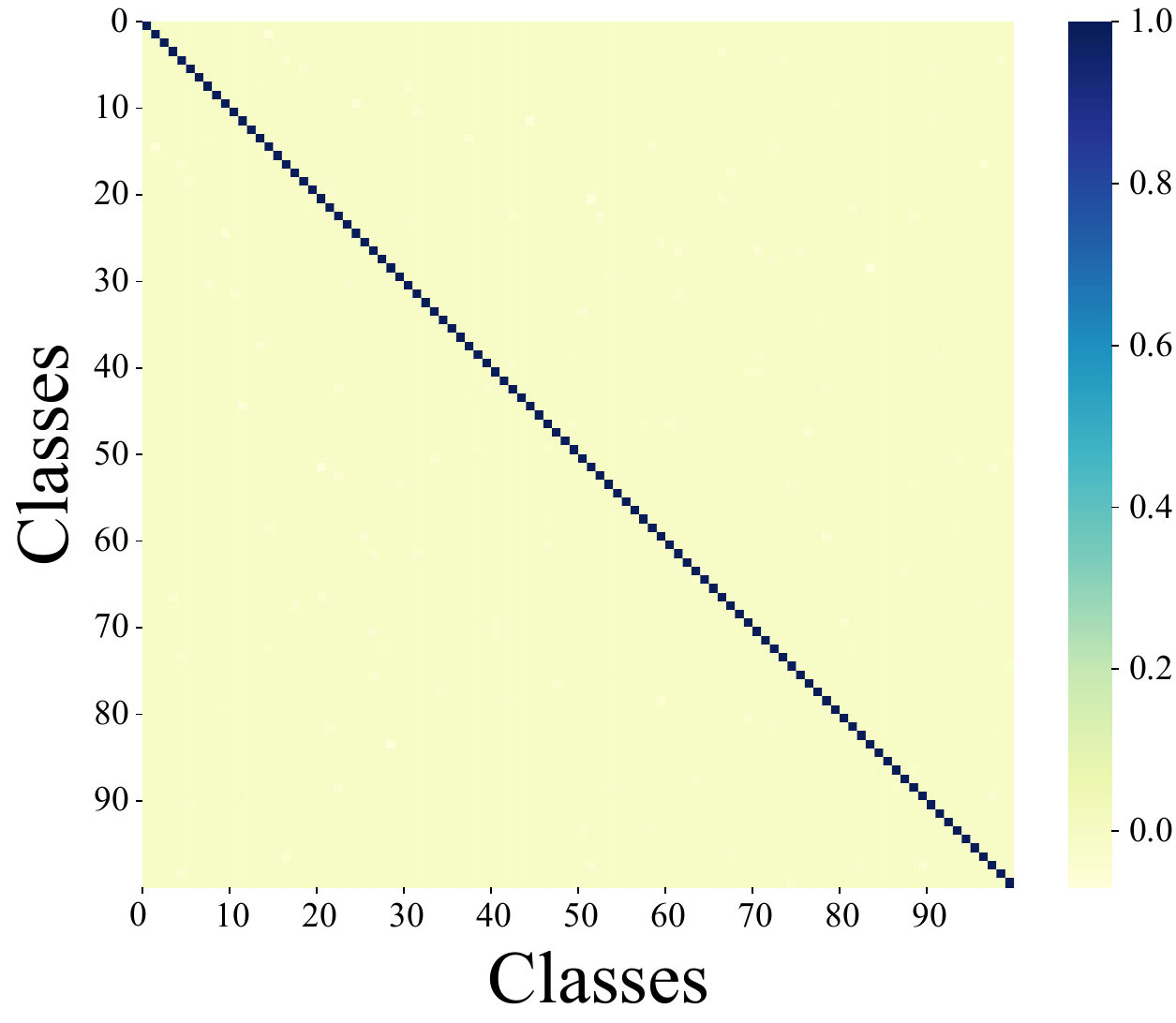}
    \caption{The heatmap of cosine similarity between classifier vectors in SSE-C.}
    \label{fig:heatmap_cosine_similarity}
\end{figure}

We opted for a learning rate of 0.0001 with 10,000 optimization steps on ImageNet-LT with 1000 classes. The order of magnitude for the norm variance is between $10^{-7}$ and $10^{-9}$. For the angles across each pair of classifier vectors, we visualized the cosine similarity in SSE-C in Fig.~\ref{fig:heatmap_cosine_similarity}, indicating that our optimized SSE-C is almost identical to the dense ETF.

\end{document}